\documentclass[conference]{IEEEtran}
\IEEEoverridecommandlockouts
\usepackage[utf8]{inputenc}

\usepackage[nocompress]{cite}
\usepackage{amsmath,amssymb,amsfonts}
\usepackage{algorithmic}
\usepackage{graphicx}
\usepackage{textcomp}
\usepackage{xcolor}

\usepackage{amsthm}
\usepackage{subcaption}
\usepackage{nicefrac}       
\usepackage{microtype}
\usepackage{booktabs}       
\usepackage{caption}
\theoremstyle{definition}
\usepackage{dashbox}
\usepackage[
    n,
    operators,
    advantage,
    sets,
    adversary,
    landau,
    probability,
    notions,    
    logic,
    ff,
    mm,
    primitives,
    events,
    complexity,
    asymptotics,
    keys]{cryptocode}
\usepackage{amssymb}
\usepackage{caption}
\usepackage{cleveref}
\usepackage{multirow}

\newtheorem{hyp}{Hypothesis}

\newcommand\shortsection[1]{\vspace{6pt}{\noindent\bf #1.}}

\newcommand{\smallfigscale}{0.367\textwidth}

\begin{document}

  
\title{\huge Subject Membership Inference Attacks in Federated Learning
}

\author{
  \IEEEauthorblockN{Anshuman Suri}
  \IEEEauthorblockA{\textit{University of Virginia}\\
    anshuman@virginia.edu}
  \and
  \IEEEauthorblockN{Pallika Kanani}
  \IEEEauthorblockA{\textit{Oracle Labs}\\
    pallika.kanani@oracle.com}
  \and
  \IEEEauthorblockN{Virendra J. Marathe}
  \IEEEauthorblockA{\textit{Oracle Labs}\\
    virendra.marathe@oracle.com}
  \and
  \IEEEauthorblockN{Daniel Peterson}
  \IEEEauthorblockA{\textit{Amazon}\\
    dwpete@amazon.com}
}
\maketitle

\thispagestyle{plain}
\pagestyle{plain}

\begin{abstract}
    Privacy attacks on Machine Learning (ML) models often focus on inferring the existence of particular data points in the training data. However, what the adversary really wants to know is if a particular \emph{individual}'s (\emph{subject}'s) data was included during training. In such scenarios, the adversary is more likely to have access to the distribution of a particular subject than actual records. Furthermore, in settings like cross-silo Federated Learning (FL), a subject's data can be embodied by multiple data records that are spread across multiple organizations. Nearly all of the existing private FL literature is dedicated to studying privacy at two granularities -- item-level (individual data records), and user-level  (participating user in the federation), neither of which apply to data subjects in cross-silo FL. This insight motivates us to shift our attention from the privacy of data records to the privacy of \emph{data subjects}, also known as subject-level privacy.
    We propose two novel black-box attacks for \emph{subject membership inference}, of which one assumes access to a model after each training round.
    Using these attacks, we estimate subject membership inference risk on real-world data for single-party models as well as FL scenarios. We find our attacks to be extremely potent, even without access to exact training records, and using the knowledge of membership for a handful of subjects.
    To better understand the various factors that may influence subject privacy risk in cross-silo FL settings, we systematically generate several hundred synthetic federation configurations, varying properties of the data, model design and training, and the federation itself. Finally, we investigate the effectiveness of Differential Privacy in mitigating this threat.
\end{abstract}


\maketitle

\thispagestyle{plain}
\pagestyle{plain}

\section{Introduction} \label{sec:intro}

Membership inference attacks on Machine Learning (ML) models aim to infer the presence of a given data record in the data used to train the models~\cite{shokri2017membership}.  Most often, this corresponds to testing for membership of certain train samples, like health records, in the training data of ML models, given access to just the fully trained models. This threat model has received a lot of attention in the ML privacy community, with numerous works on formalizing and theorizing this inference risk~\cite{dwork06a, yeom2018privacy}, proposing attacks~\cite{shokri2017membership, jayaraman2021revisiting}, and mitigation strategies deployed in practical scenarios~\cite{abowd20222020, thakurta2017learning}. However, the membership inference threat model hinges on one critical assumption: the adversary has access to exact records potentially part of training data. Studying membership inference is useful for auditing and to test model memorization, but remains disconnected from goals of actual adversaries that care about targeting individual persons and their data, and not necessarily particular data records. In realistic settings, the adversary often does not even have access to the exact data records used for model training.

Consider an adversary that wants to probe a ML model to test membership of an individual's data in the model's training data. An adversary is more likely to have access to some representative face images of the target individual (also referred to as \emph{subject} or \emph{data subject}), but not necessarily the ones used for training the model~\cite{chen2023face}. A membership inference attack is thus ill-defined in such a setting, since access to exact records is unavailable.
This threat model, where the attacker is interested in identifying existence of a subject in the training data without necessarily using actual training data records, can be  termed as \emph{subject membership inference}, and serves as a counterpart to subject-level privacy~\cite{marathesubject}. 

Subject membership inference can be thought of as a subset of distribution inference~\cite{suri2022formalizing}: the adversary seeks to infer whether the training data was sampled from a distribution with the target subject's data ($\mathcal{D}_1$) or not ($\mathcal{D}_0$). \textbf{Subject-level privacy is crucial because ultimately, we are interested in preserving the privacy of an individual, not just that of a data item.}
Although existing membership-inference attacks may yield non-trivial advantage in this scenario, there is a need for attacks designed specifically to infer subject membership.
Many real-world ML models are trained on data aggregated from multiple devices, regions, or collection strategies, leading to more data per subject. This, in turn, increases the risk to these subjects' privacy. Subject membership inference attacks provide an important empirical evaluation of this risk.

\begin{figure}[t]
    \centering
    \includegraphics[trim={4.0cm 0.8cm 3.5cm 0.5cm},clip, width=0.45\textwidth]{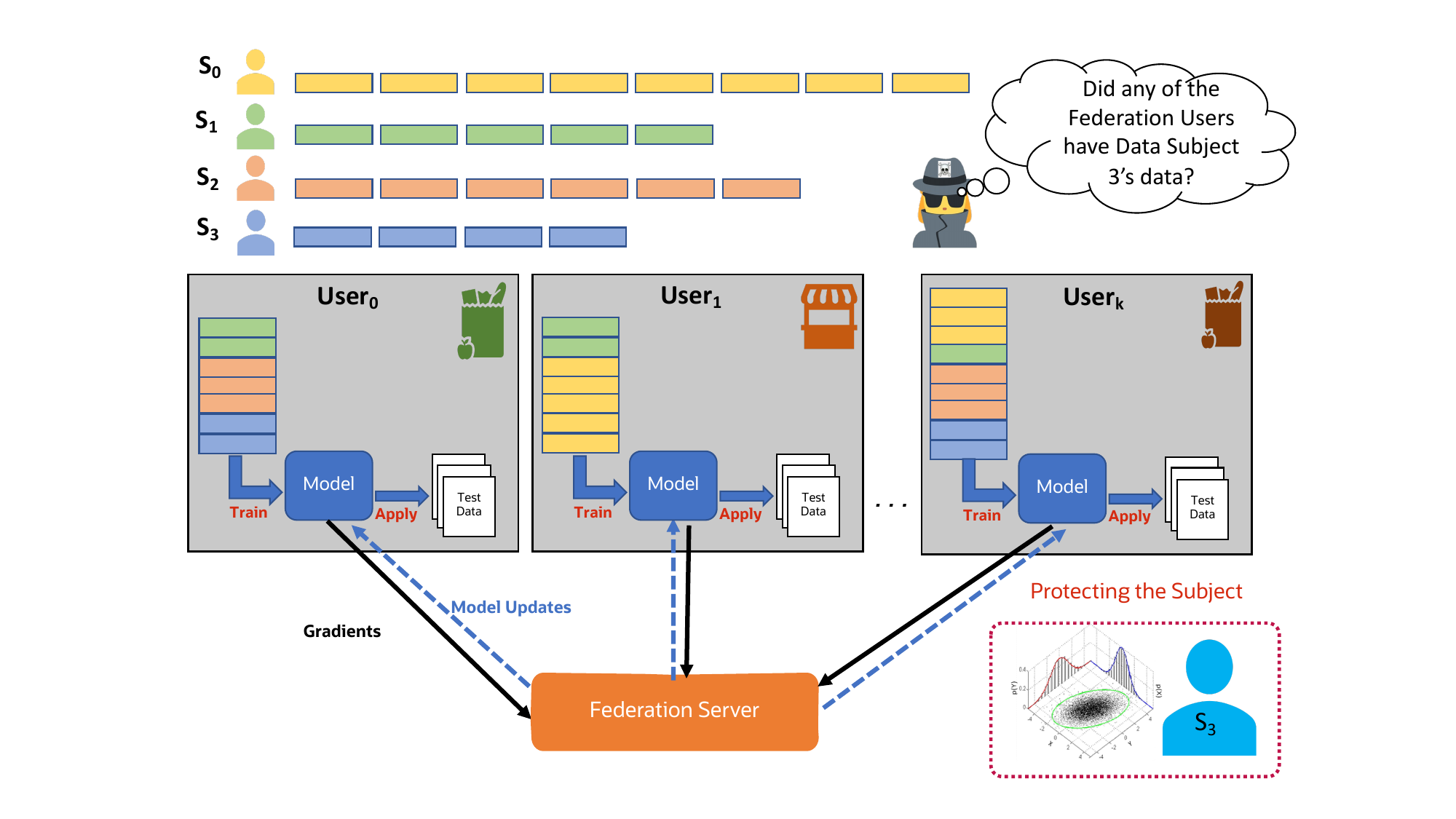}
    \caption{Subject-level Privacy in Cross-Silo Federated Learning. Data records of each the subject $S_i$ are scattered across multiple federation users.}
    \label{fig:subject-privacy}
\end{figure}

Federated Learning (FL) \cite{mcmahan2017communication} has emerged as an important paradigm that allows model training without physically sharing data. It was originally introduced for mobile devices, but is now also deployed in collaboration between large organizations or data centers across geographies, the so-called \emph{cross-silo} setting~\cite{kairouz2021advances}. The `users' of the federation in this setting are typically organizations, such as a group of retailers or hospitals, who in turn are collecting data from a large number of individuals, and an individual data subject can have their data records distributed among different federation users (e.g. a patient's health records at different hospitals, a shopper's purchase history at several retailers, internet data aggregators). Figure~\ref{fig:subject-privacy} depicts such a cross-silo FL setting where data records of subjects $S_i$ are distributed across multiple federation users.

In the \emph{cross-device} FL setting, privacy is usually defined at two \emph{granularities}: \emph{item-level privacy}, which describes the protection of individual data items~\cite{abadi16,mcmahan18} and \emph{user-level privacy}, which describes the protection of the entire data distribution of the user device~\cite{liu2020learning,mcmahan18}.
Item-level privacy can also be regarded as a specific instance of subject-level privacy in FL, where a subject's data manifests as a single data record at a single federation user. For user-level privacy, it may be tempting, and mostly accurate, to draw an equivalence between subject-level privacy and user-level privacy, from the perspective of cross-device FL setting.  For instance, a federation of cell phones has a one-to-one mapping between data subjects (each individual owning a participating cell phone) and federation users (the cell phones). However, this equivalence breaks down in cross-silo settings, where an individual's data is spread across several federation users, or organizations. Recently, subject-level privacy ~\cite{marathesubject} has been formalized as a distinct granularity.

The privacy risks associated with users and subjects can vary wildly in such settings.
In fact, it is trivial to construct a situation where each federation user has an identical data distribution despite vastly different subject-level distributions, simply by allocating each subject's data evenly across the federation servers. For example, consider retailers as federation users and shoppers as data subjects. Different shoppers with very different shopping habits might still lead to overall uniform distribution for each retailer, as long as their data is evenly distributed across retailers. Similarly, if each federation user collects a biased subsample of subject data, then identical data subject distributions may yield federation users with extreme heterogeneity. In the retailer example, this is equivalent to shoppers who have similar purchase patterns, but it might be unevenly distributed across retailers. 

\begin{figure}
\centering
    \centering
    \includegraphics[trim={0.5cm 0.15cm 1.6cm 1.4cm},clip,width=\smallfigscale]{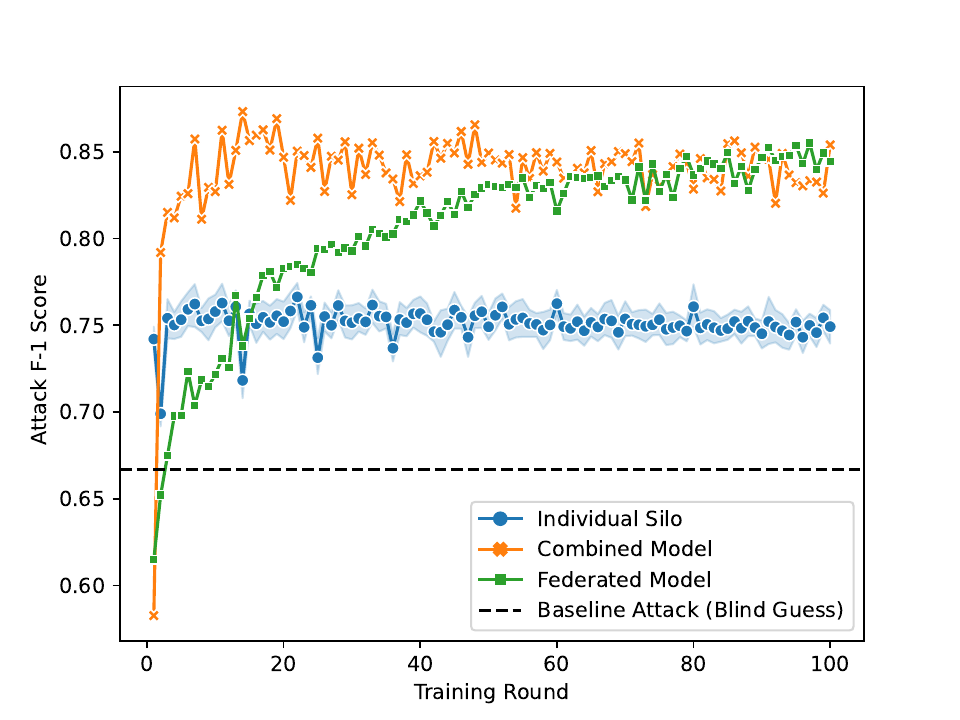}
\caption{Subject Membership inference risk (F$_1$ score) across training rounds for the average of individual silos (blue), models trained on data combined from all users (orange), and models trained with FL (green), on the FEMNIST dataset (details in Section~\ref{sec:exp_setup}). Aggregation of data increases risk significantly, and FL is just as bad as physically combining data for the trained models.}
\label{inference_risk_fig}
\end{figure}

Although FL aims to protect data privacy of participants by restraining individual data items to their respective local sites, it does not protect the privacy of subjects, whose data can be scattered across participants. To illustrate this, we simulate a cross-silo federation, in which many subjects have their individual data points spread across several silos. We first train individual classifiers on each of the silos, and measure the subject membership inference risk. As we can see from Figure~\ref{inference_risk_fig}, aggregating data for subjects to train a single joint model increases privacy risk, regardless of whether the combination happens by physically moving the data to a central server or through Federated Learning.

The real-world implications of this finding cannot be overstated. Despite a growing number of regulations regarding data privacy being proposed globally, data subjects often have limited control over how their data gets used by the external entities that they interact with. When these entities, such as banks, hospitals, retailers, choose to participate in federations to build more accurate ML models, they might inadvertently put the data subjects' privacy at risk. Subject membership inference attacks would be the primary tools necessary for accurately estimating this risk. 

The success of privacy attacks on ML models depends on both the nature of the training data and the type of modeling technique. A FL system with multiple users and data subjects can be quite complex, and various factors can greatly influence the effectiveness of privacy attacks. Studying trends in inference risk with variations in these parameters is thus equally important, as certain configurations may be more susceptible than others. Knowing where on that spectrum does a particular system configuration lies, along with an estimate of the associated risks case enable a practitioner make informed decisions about the system design choices.

\shortsection{Contributions}
We propose two new attacks for subject-level membership inference that require only black-box access to model predictions (Section~\ref{sec:method}) and partial knowledge about subjects whose data was used in training. One of these attacks applies to trained ML models, while the other assumes access to intermediate model states across training and is more relevant when the adversary is a participant in the federation.
Experiments with a real-world dataset, FEMNIST~\cite{caldas2018leaf}, suggest that these attacks are very effective in accurately predicting subject membership inference whether the attacker has access to samples from the actual training set, or only representative samples from the subjects' data distribution (Section~\ref{sec:sub_vs_mi_inf}). Attack inference F$_1$ scores are as high as $0.8$ when subject membership is known for as few as 5 subjects. 

Next, we assess the effectiveness of a popular mitigation strategy prescribed for ML privacy -- Differential Privacy (DP)~\cite{dwork2006calibrating} in providing protection against these attacks.
We retrain models on FEMNIST with DP at the granularities of data items~\cite{dwork06a}, federation users~\cite{abadi16}, and data subjects~\cite{marathesubject}, and repeat the attacks. We confirm that DP is indeed effective in reducing privacy leakage, but hurts task accuracy.
We also show experiments on the Shakespeare dataset \cite{caldas2018leaf}, which is an example of a dataset where subject distributions are almost identical to each other, decreasing attack effectiveness.

To more broadly understand the impact of FL configurations on the  subject membership inference risk, we design a systematic and thorough evaluation of the proposed attacks.  We begin with building a synthetic data simulator, capable of simulating different federation configurations (Section~\ref{sec:experiments}). Each configuration consists of multiple user datasets, which in turn are composed of data from multiple data subjects.
Our evaluations across these configurations reveal trends with respect to data, model, and federation properties, providing actionable guidance to practitioners on assessing and mitigating subject membership inference risks (Section~\ref{sec:conclusion}).

\section{Background}
\label{sec:background}

Our work is concerned with what adversaries can learn about training data given access to machine learning models, which may constitute a breach of a data subject's privacy. There are many attack surfaces for a machine learning model across its life-cycle, including after deployment~\cite{Truong2021-cd,jegorova2022survey,liu2022ml}.

Privacy attacks are a common approach used to assess privacy risks in ML. The key benefit of this approach is that it grounds the privacy discussion concretely in terms of the training data whose privacy can be compromised by the model.
These privacy attacks can include threats such as model inversion~\cite{fredrikson2015model, zhang2020secret} and property inference~\cite{ateniese2015hacking, ganju2018property, suri2022formalizing}. However, the majority of research has focused on membership inference attacks~\cite{shokri2017membership, yeom2018privacy}.
Membership inference is a popular type of privacy attack that is highly relevant to our work: determining whether a particular data item was part of the training dataset. A successful membership inference attack concretely demonstrates privacy risks to individual data items from the attacked model's training dataset.

We particularly follow the line of prior work on membership inference attacks~\cite{Shokri2017-tg,Salem2019-pk,Shejwalkar_undated-lk}, which deliver an empirical lower bound to the risk of data leakage through a ML model. The adversary’s accuracy in determining whether a data point is part of the training set gives a very real picture of whether the data was leaked.

Several advancements in this line of work improve this lower bound by making attack models more accurate and applicable in more realistic scenarios~\cite{jayaraman2021revisiting}. In particular, white-box attacks that rely on the gradients sent during training show that these gradients reveal a lot of detail about the training data~\cite{zhu2019deep,mo2020layer,Geiping2020-nn,geng2023improved}, although the attacks may be defended against with appropriate strategies~\cite{huang2021evaluating}. White-box attacks are quite plausible if the adversary is posing as a legitimate user in federated learning, and this opens up new avenues of risk~\cite{Wainakh2021-io,Liu2021-qo,Hitaj2017-tq} that are highly relevant to our work.

While these advanced attacks are worth investigating to improve subject-level membership inference, they are overwhelmingly focused on membership inference of particular data points~\cite{nguyen2023active}. User-level privacy leakage~\cite{Wang2019-qy} is more closely tied to subject-level privacy leakage that we study here, and indeed is equivalent under settings with a one-to-one correspondence between subjects and federation users. Additional recent work on membership inference attacks against word embeddings, text classifiers, and language models attempts to determine whether an author's text was part of the training dataset~\cite{mahloujifar2021membership}. This work exactly fits our notion of subject-level membership inference, and their attacks cleverly use subsets of bigrams to distinguish author membership, but at the cost of being very domain-specific, and their attacks incur a much higher computational cost than the attacks presented in this work.
More recently, privacy leakage FL has received considerable attention, with works ranging from property-inference poisoning attacks~\cite{wang2022poisoning} to passive adversaries inferring gender from voice samples~\cite{tan2023general}.

One of the main contributions of our paper is that we empirically test the success of inference attacks as properties of the data distribution change. This is related to prior work that examines patterns of membership inference success as the architecture of the model changes~\cite{Truex2019-iz} across an even wider array of model architectures, but which focuses on item-level membership inference and a smaller number of datasets.

\section{Subject Membership Inference Attacks }
\label{sec:attacks}

We begin by formally describing the adversary's objective (Section~\ref{sec:attacker_objective}), followed by a description of the threat model and assumptions about data and model access. Our attacks require only black-box API access, and are thus applicable to any ML model. One of them only assumes access to the final trained model (Section~\ref{sec:threat_model_data}), while the other assumes access after each training round, and is thus more suited to FL settings (Section~\ref{sec:threat_model_fed}).

\subsection{Attack Objective} \label{sec:attacker_objective}

Let $\mathcal{S}_0$ be a set of subjects, and $s_\text{interest}$ the subject whose membership the adversary wants to infer, such that $s_\text{interest} \not\in \mathcal{S}_0$. Let $\mathcal{D}^s$ be the distribution corresponding to a subject $s$. Then, using the definitions of distribution inference in~\cite{suri2022formalizing}, we can formulate our subject membership inference task as differentiating between models trained on datasets sampled from either of the distributions $\mathcal{D}_0$ and $\mathcal{D}_1$, defined as:
\begin{align}
    \mathcal{D}_b = \bigcup_{s \in \mathcal{S}_b} \mathcal{D}^s
\end{align}
where $\mathcal{S}_1 = \mathcal{S}_0 \cup \{s_\text{interest}\}$. This is equivalent to stating that a data sample from either of $\mathcal{D}_{\{0,1\}}$ is equivalent to taking a union of samples from the individual subjects' distributions. The first distribution $\mathcal{D}_0$ corresponds to the absence of subject of interest in the federation, while $\mathcal{D}_1$ includes it.
\textbf{A Subject membership inference attack thus aims to infer whether a given subject's data was used in the federation, \textit{i.e.} was present in \textit{any} of the users' datasets.} The flow of information for the proposed subject membership inference attack is described in Figure~\ref{fig:information-flow}.

Note that subject membership inference is orthogonal to the FL setting, and is indeed more broadly applicable to ML models. For subject membership inference in FL, it is important to note that it does not matter how a subject's data is divided across different users of the federation. Even if only one user has the subject's data, or if an individual subject's data is divided across all users, the subject's data is ultimately used in the overall training process and thus the subject should be inferred as being present. The adversary only cares about the subject's presence in the overall federation and using the above formulation is apt for the given threat model.

\subsection{Threat Model} \label{sec:threat_model}

Attacks may be grouped by whether or not the adversary has knowledge (or perhaps partial knowledge) of the model, into black-box, white-box, and grey-box attacks~\cite{nasr19,Truex2019-iz}. Participants in FL have access to the model architecture and parameters, giving white-box access to all members of the federation. We do not assume the attacker is part of such an FL setup, or that it has white-box access to the victim's model. Our Loss-Threshold Attack (Section~\ref{sec:membership_inference}) uses only knowledge of the data points, the labels assigned by the model, and the loss function the model is optimizing; which is essentially a black-box attack.
The Loss-Across-Rounds Attack (Section~\ref{sec:membership_inference}) additionally assumes API access to the model after each training round, and is designed to extract additional information from the model's training behavior across time. This attacker can exist as an honest-but-curious federation server/user in the federation. In either case, by design the attacker has access to the global model after each training round.
For all of our attacks, we assume the adversary has access to the following:

\begin{figure*}[]
    \centering
    \includegraphics[trim={0.7cm 4.7cm 0.7cm 4.7cm},clip, width=0.7\textwidth]{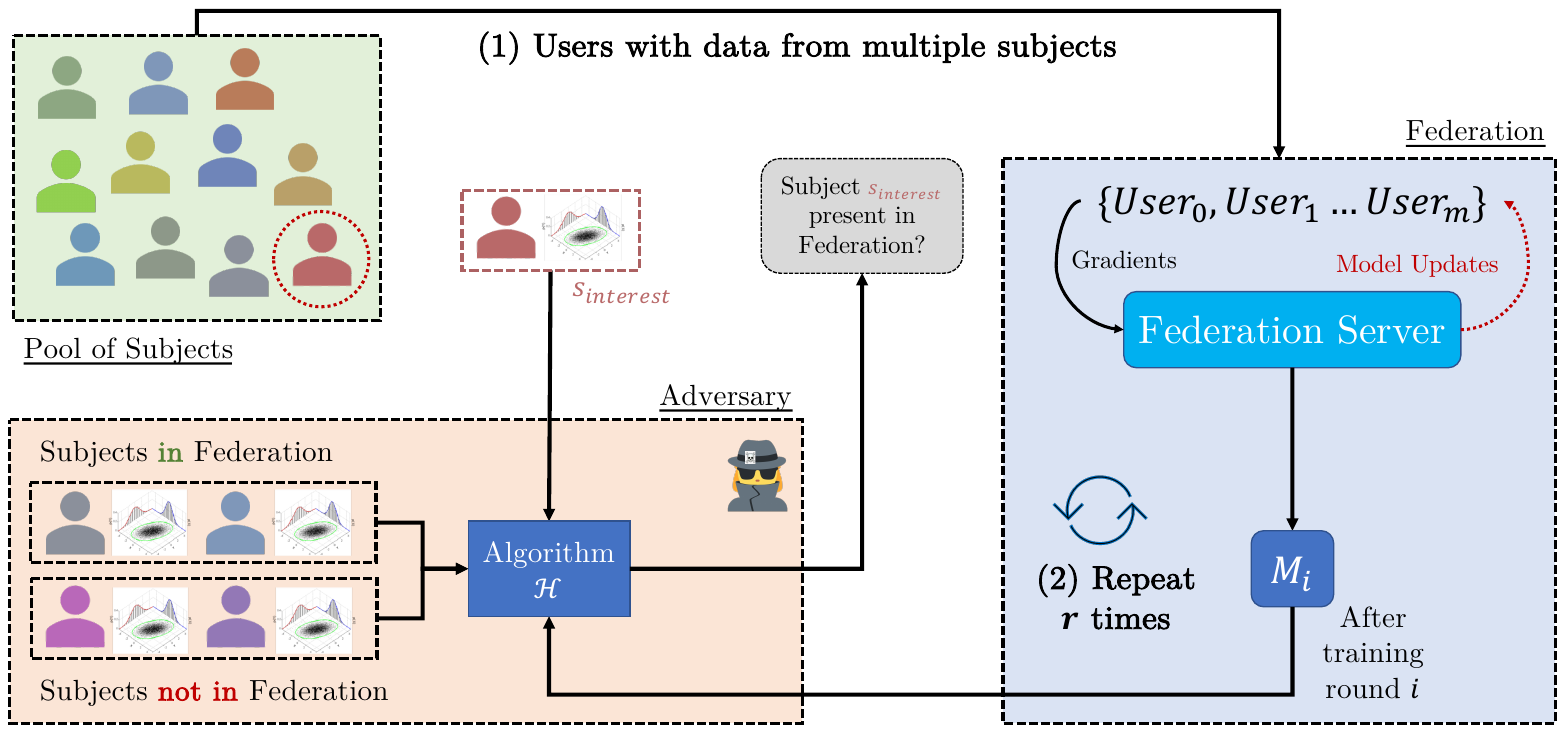}
    \caption{Information flow for the subject membership inference attack in Federated Learning. The adversary uses some algorithm $\mathcal{H}$, along with knowledge of membership of a few subjects and the models $M_i$ after each training round $i$, to infer the membership of $s_\text{interest}$'s data in any of the user's data.
    }
    \label{fig:information-flow}
\end{figure*}

\subsubsection{Samples (finite) from the distribution of subjects} \label{sec:threat_model_data}
If the adversary wishes to launch an attack against a particular subject, it must have the capability to quantify and differentiate subjects and identify the one it is interested in. This can be done by either knowing (or estimating) a subject's distribution or possessing finite samples to estimate it. Having access to finite set of samples from the subject's distribution is the weaker assumption of these two. Note that in theory, it is not necessary to have estimates for distributions for \emph{all} of the subjects--- just for the subject of interest, and some samples from subjects with known inclusion/exclusion labels. In the first paper on membership inference, these subjects with known labels formed a training set for a machine learning model~\cite{shokri17}, but the attack models in this work compare the model's loss against simple thresholds. We assume the attacker has access to a limited number of known included/excluded subjects, and samples from each distribution, in order to tune the threshold values.
Since our attacks do not require training any shadow models (which is the case for most state-of-the-art membership~\cite{ye2022enhanced}, distribution inference~\cite{suri2023dissecting} or subject-membership~\cite{chen2023face} attacks), it is general and can be applied in very-low-data settings.

\subsubsection{API access to the global model after each federation round} \label{sec:threat_model_fed}
We assume access to prediction probabilities from the global model after each training round. Both the central server and individual participants have access to the global model after each training round, making it easy to satisfy this requirement. This may be further weakened to limit access to just the last round- the final global model that may be released to the world. We thus propose two attacks; one for each level of access described here.

\section{Method}
\label{sec:method}

Both of our attacks are based on hypotheses implied by prior works on the behavior of loss functions on training data~\cite{yeom2018privacy,jayaraman2021revisiting, ruder2016overview}.
Given the objective of training ML models, it is natural to expect that the model's performance on data similar to that seen during training would be better than that not seen during training.
The Loss-Threshold Attack (Section~\ref{sec:membership_inference}) is generic and applicable to any ML model with black-box access, while the Loss-Across-Rounds Attack (Section~\ref{sec:loss_across_rounds}) assumes access to intermediate model state during training, and is thus more suitable for FL settings.

Let $m$ be the total number of users participating in the federation. Let $r$ be the number of rounds for which the global model is trained in the federation, with $M_i$ denoting the state of the model after training round $i$ has completed. $M_0$ thus represents the state of the model before training starts. Let $l_i(x,y)$ be the loss value between the label $y$ and $M_i(x)$, with $M_i(x)$ denoting the model $M_i$'s prediction on point $x$.

\subsection{Loss-Threshold Attack} \label{sec:membership_inference}
\hyp{\textit{If data from a particular subject is present in the federation and is used in training, the global model would be expected to have a lower loss on it than data from a subject that was not present in any of the users' local datasets~\cite{yeom2018privacy}.}}

Based on this hypothesis, we propose the following attack: record loss values for samples from the target subject's distribution and check if any of them have a value less than a particular threshold. If the loss is below the threshold, it would indicate the model has seen that particular data, and thus other data from that subject's distribution, during training.
\begin{align}
    c & = \sum_{(d_x, d_y) \sim \mathcal{D}_s}\mathbb{I}[l_r(d_{x}, d_{y}) \leq \lambda]
    \label{equation:loss_threshold}
\end{align}
The adversary can either check if $c$ is non-zero or derive an additional threshold on this value based on the metric it wishes to maximize, like precision or recall.

\subsection{Loss-Across-Rounds Attack}
\label{sec:loss_across_rounds}
\hyp{\textit{Loss on training data, and thus data from the training distribution, decreases across iterations by virtue of how learning algorithms  work. However, data from distributions not seen during training would likely not converge to values as low as those of subjects present in the federation~\cite{wanggeneralizing}.}}

Based on this hypothesis, we propose the following attack: record loss values for samples from the subject's distribution and note how the loss values change as training rounds progress. The attack first computes the loss across each training round $i$:
\begin{align}
    c_i = \sum_{(d_x, d_y) \sim \mathcal{D}_s}l_i(d_x, d_y)
\end{align}
Then, the adversary takes note of the number of training rounds where the loss decreases after each round:
\begin{align}
    c = \sum_{i=1}^r \mathbb{I}[c_i < c_{i-1}]
\end{align}
The adversary can then compute these values for both subjects seen and not seen in the federation and consequently derive a threshold on this value for subject membership. The attack implicitly assumes all users contribute in each training round. Although this assumption may not hold in most settings, the likelihood of any user chosen in a training round containing a subject's data is non-trivial. This, coupled with the robustness of learning algorithms over individual rounds~\cite{hardt2016train}, is sufficient to launch the attack and achieve high inference leakage.

\subsection{Threshold Tuning}
\label{sec:threshold_tune}
Membership inference attacks label whether a subject is part of the training data; it is common for these labeling strategies to depend on some
parameters or hyper-parameters, like any ML system. These hyper-parameters are usually computed using additional information that may be available through side-channel attacks or just by the adversary participating in the federation training.

All of our attacks involve computing some form of tunable thresholds that are used to execute the attacks (e.g., $\lambda$ in Equation~\ref{equation:loss_threshold}). The threshold values affect the precision/recall tradeoff of the attack, and in this work we learn their optimal values from a data set of correctly labeled included and excluded subjects.
For the scenario where the adversary is a participant in the federation, it can use its split of data to generate a training set for the attack. At the very least, the attacker knows for certain that subjects for which records exist in its training data are part of the federation. 
The adversary can then guess subjects that are likely not used in the federation by randomly sampling (or generating) other subjects not in their data, or by intentionally holding some data out from the training of the federated model. 
Once data for both subjects used and (probably) not used during training is available, the adversary can tune the thresholds of their chosen attack to accurately predict whether a subject's data was used in the federation or not.

\section{Evaluation}
\label{sec:sub_vs_mi_inf}

Equipped with knowledge of the threat model and the adversary's capabilities, we  move on to the following questions:
\begin{enumerate}
    \item \emph{How well can an adversary perform in the absence of access to exact data from training, while testing for subject membership inference?}
    \item \emph {How little information about subject inclusion/exclusion can the adversary get away with while still being effective? }
    \item \emph{How does FL affect subject membership inference risk, compared to standard  training?}
    \item \emph{How well do these attacks hold up against differential-privacy based mitigation approaches?}
    \item \emph{How do the properties of the data, model and federation affect attack performance?}
\end{enumerate}

To answer these questions, we use a real-world dataset and train models on it with both standard and FL training (Section~\ref{sec:exp_setup}). We then test out the efficacy of our attacks under various training environments and defenses (Section~\ref{sec:results}). Our evaluations reveal how distribution-based attacks can be just as potent as ones based on exact record membership, both in extracting subject membership information, and evading defenses (Section~\ref{sec:dp_effect}).

\subsection{FEMNIST Experimental Setup} \label{sec:exp_setup}

We use FEMNIST~\cite{caldas18}, the federated extended MNIST~\cite{deng2012mnist} dataset, an image classification task for handwritten digits and letters. FEMNIST's digits and letters themselves have been written by $3500$ distinct individuals, and FEMNIST partitions these images by individual authors. Each author has contributed hundreds of sample images. Ordinarily, FL research experiments~\cite{caldas18} map each author to a federation user, resulting in a $3,500$-user federation. In our experiments, we instead map authors to subjects, and reserve half of the subjects as non-member subjects, resulting in a federation with $1750$ subjects whose data are randomly scattered among a handful of federation users ($16$ in our experiments). The remaining half ($1750$) subjects are never involved in the federation and are ``non-members". Each subject has $\sim140$ data points on average, with its data more-or-less equally spread across $16$ federation users. Multiple federation users may host images from the same subject, though we do not distribute any individual image to more than one federation user. This reconfigured dataset is especially suitable for cross-silo FL and our subject membership attacks study. The data points themselves are $28$x$28$ pixel, black-and-white pictures of a single handwritten digit or letter.

\begin{figure}[ht]
\centering
\begin{subfigure}[b]{\smallfigscale}
    \centering
    \includegraphics[trim={0.5cm 0.1cm 1.6cm 1.3cm},clip,width=\textwidth]{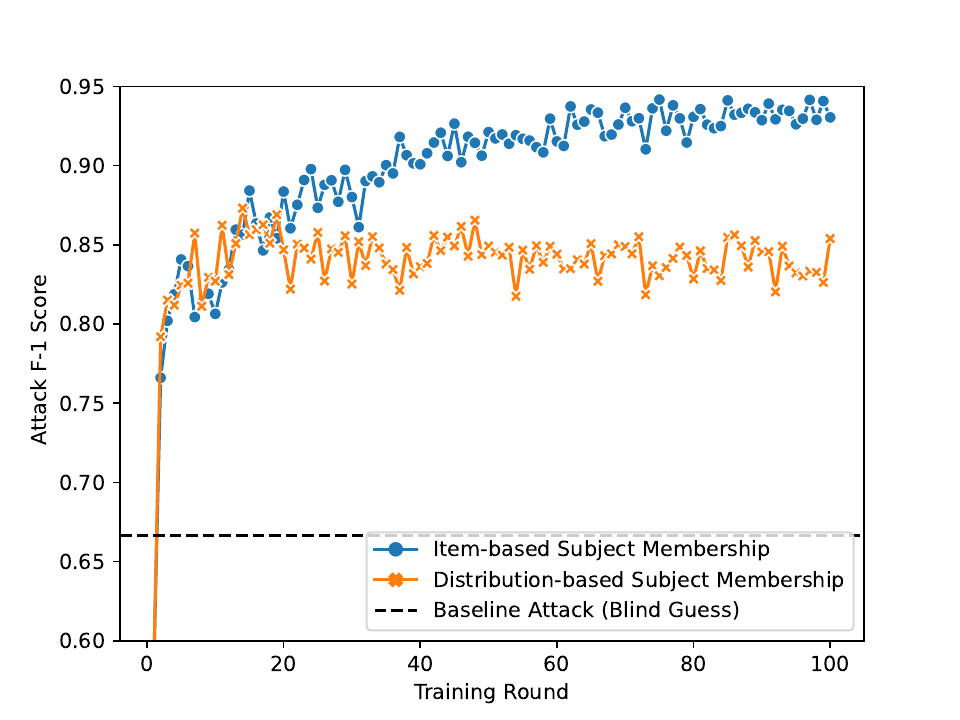}
    \caption{Normal Training}
    \label{fig:mi_vs_si_normal}
\end{subfigure}
\hfill
\begin{subfigure}[b]{\smallfigscale}
    \centering
    \includegraphics[trim={0.5cm 0.1cm 1.6cm 1.3cm},clip,width=\columnwidth]{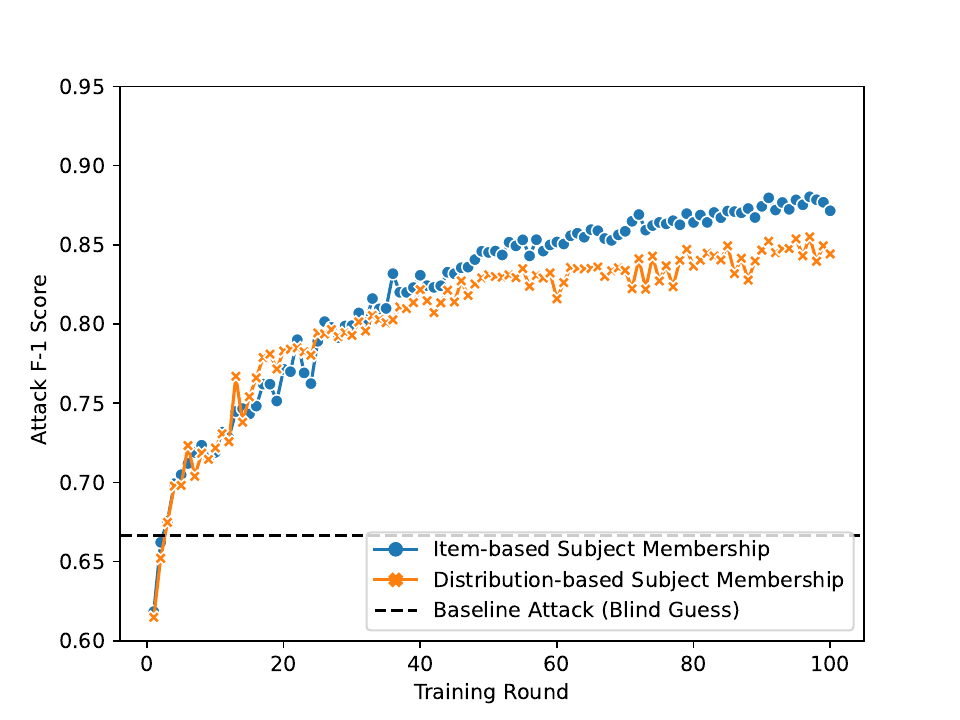}
    \caption{Federated Learning}
    \label{fig:mi_vs_si_fl}
\end{subfigure}
\caption{Attack F-1 Score across training rounds inferring subject membership using our attack (orange) and via membership inference attacks using exact records (blue), for normal training (\protect{\subref{fig:mi_vs_si_normal}}) and FL (\protect{\subref{fig:mi_vs_si_fl}}). Inference risk is not harmed significantly by the lack of access to exact data, and can be just as bad for FL as normal training.
}
\label{fig:mi_vs_si}
\end{figure}

\shortsection{Target Model Training}
We use the CNN model on FEMNIST appearing in the LEAF data suite~\cite{caldas18} as our target model to train. More specifically, the model consists of two Convolution layers interleaved with ReLU activations and Max-pooling layers, followed by two Linear layers.
We use half of the data points belonging to each of the $1750$ member subjects for training, and reserve the remaining half as in-distribution data points to sample from for carrying out the distribution-based attack. In the standard model, we simply combine the data from all member subjects to train a standalone model using Stochastic Gradient Descent, while for the federated setting, we use FedAvg\cite{kairouz2021advances} training protocol. We train each model for $100$ rounds using the Adam\cite{kingma2014adam} optimizer, with a learning rate of $0.001$ and batch size $512$. 

\shortsection{Attack Set, Threshold Tuning, and Evaluation} We prepare the attack set by sampling at most $100$ examples from both member (the other half of each member subject's data, as mentioned above) and non-member (included and excluded) subjects. We then split this data by \emph{subject} into two parts. The first split is used by the adversary to derive the subject membership inference threshold(s) $\lambda$ (we call this the validation set), while the second split is used for evaluating the effectiveness of the attack and reporting results. 
We compute attack F$_1$ scores to measure adversary success; we count correctly predicting the presence/absence of a subject's data in the federation as a hit (1) and incorrect prediction as a miss (0).

\subsection{Results on FEMNIST}
\label{sec:results}

We evaluate how the lack of access to exact records affects inference risk, revealing how attacks retain much of their potency (Section~\ref{sec:weaker_adv_assumptions}).  Much of this potency is retained even as the number of subjects for validation decreases. These results hold for both our attacks, which we find to be similar in performance (Section~\ref{sec:attack_compare}). Finally, we evaluate DP mechanisms at different granularities and privacy budgets as defenses (Sections\ref{sec:dp_effect}).

\subsubsection{Item based v/s Distribution based} \label{sec:weaker_adv_assumptions}
To consider the impact of not having access to exact data records, we design two versions of our attacks. The first version (\emph{Item-based}) assumes access to exact records used in model training while testing for subject membership, while the second version (\emph{Distribution-based}) only assumes access to a subject's distribution. As expected, there is a gap in performance between the two settings, as the item-based access model makes much stronger assumptions (Figure~\ref{fig:mi_vs_si_normal}). Nonetheless, attack performance with just distribution-based access is high, achieving attack F-1 scores $\geq0.85$. \textbf{An adversary can thus perform subject membership inference with high success, even without access to exact records.} Interestingly, we also note that inference risk is high after as few as two training rounds suggesting that subject membership inference, unlike membership inference~\cite{yeom2018privacy}, is high even when before the model has overfit. Next, we train ML models via Federated Learning, and repeat our attacks with the same two versions as the experiment above. Not only do we observe similar attack performance for both variants, but the difference in their efficacy is also even lower when data is aggregated via FL.

\begin{figure}[h]
\centering
\begin{subfigure}[b]{\smallfigscale}
    \centering
    \includegraphics[trim={0.5cm 0.1cm 1.6cm 1.3cm},clip,width=\textwidth]{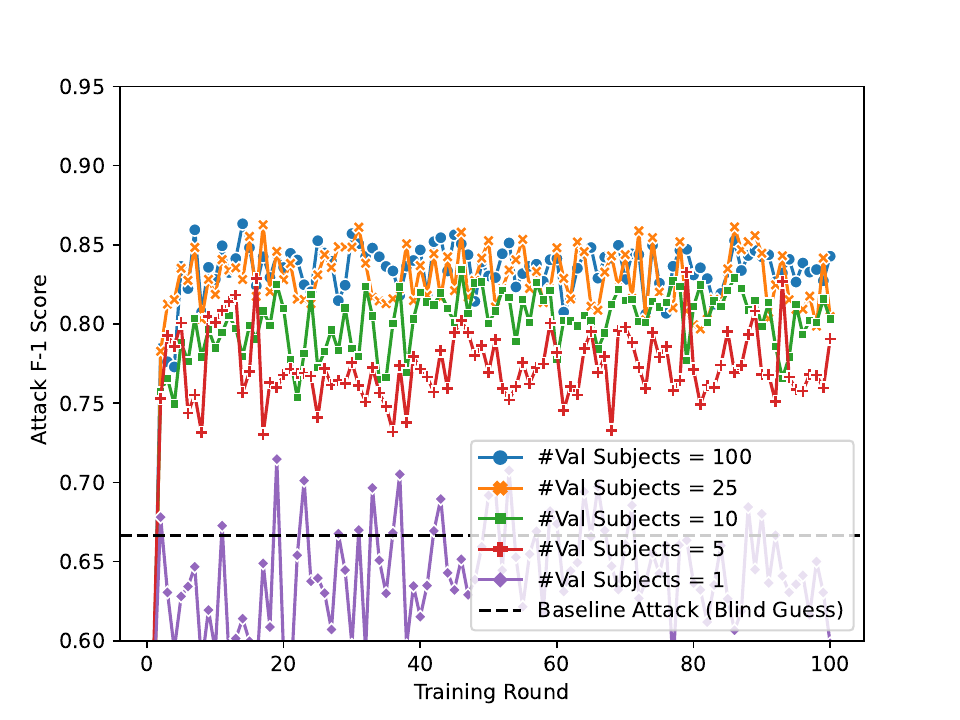}
    \caption{Normal Training}
    \label{fig:num_subject_val_normal}
\end{subfigure}
\hfill
\begin{subfigure}[b]{\smallfigscale}
    \centering
    \includegraphics[trim={0.5cm 0.1cm 1.6cm 1.3cm},clip,width=\columnwidth]{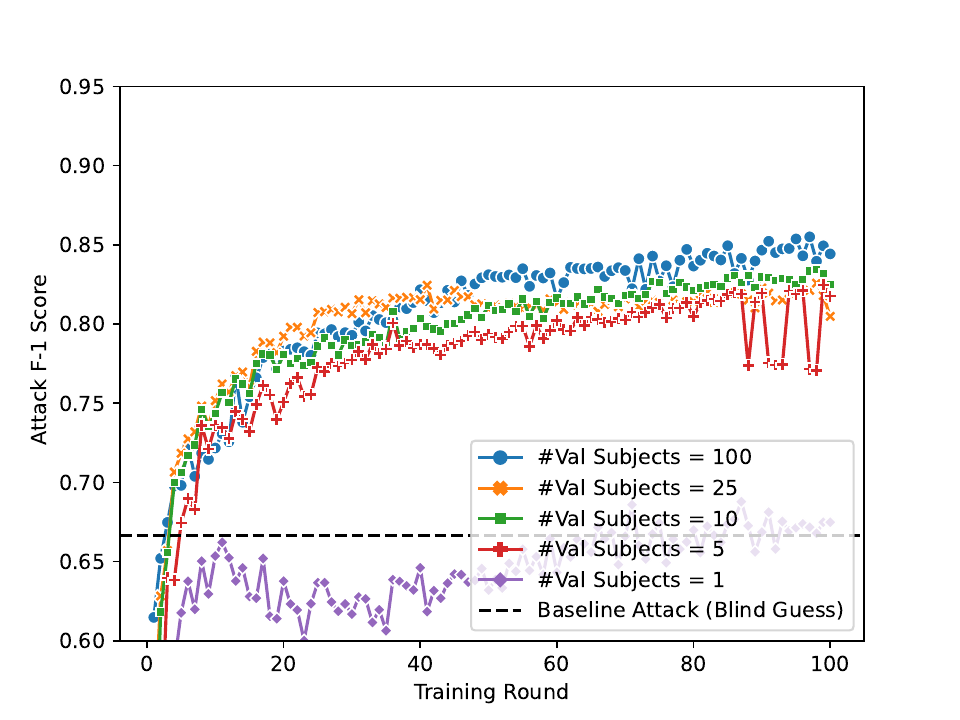}
    \caption{Federated Learning}
    \label{fig:num_subject_val_fl}
\end{subfigure}
\caption{Attack F-1 Score across training rounds inferring subject membership, for normal training (\protect{\subref{fig:num_subject_val_normal}}) and FL (\protect{\subref{fig:num_subject_val_fl}}), while varying the number of subjects in-set used for validation.
Inference risk is robust to the number of subjects used for validation, and is quite high for as few as 5 subjects. We observe similar trends for F$_1$ for the Item-level Subject Membership scenario (Figure~\ref{fig:num_subject_val_item} in the Appendix).
}
\label{fig:num_subject_val}
\end{figure}

We assume knowledge of membership for 100 subjects for computing thresholds. This is relatively small fraction ($\sim5\%$), but may be hard to obtain in some scenarios. To measure changes in inference risk as this number is lowered, we repeat experiments for both standard and FL training while varying this number in $\{1, 5, 10, 25\}$. In both settings, inference risk is near-random when only one subject's membership is known, but fairly robust as long as membership for $\geq10$ subjects is known (Figure~\ref{fig:num_subject_val}). Our evaluations demonstrate how \textbf{adversaries can be fairly successful with knowledge of as few as five subjects} ($\sim0.3\%$ of all subjects).

\subsubsection{Attack Variants} \label{sec:attack_compare}
\begin{figure}[ht]
\centering
    \centering
    \includegraphics[trim={0.5cm 0.2cm 1.6cm 1.3cm},clip,width=\smallfigscale]{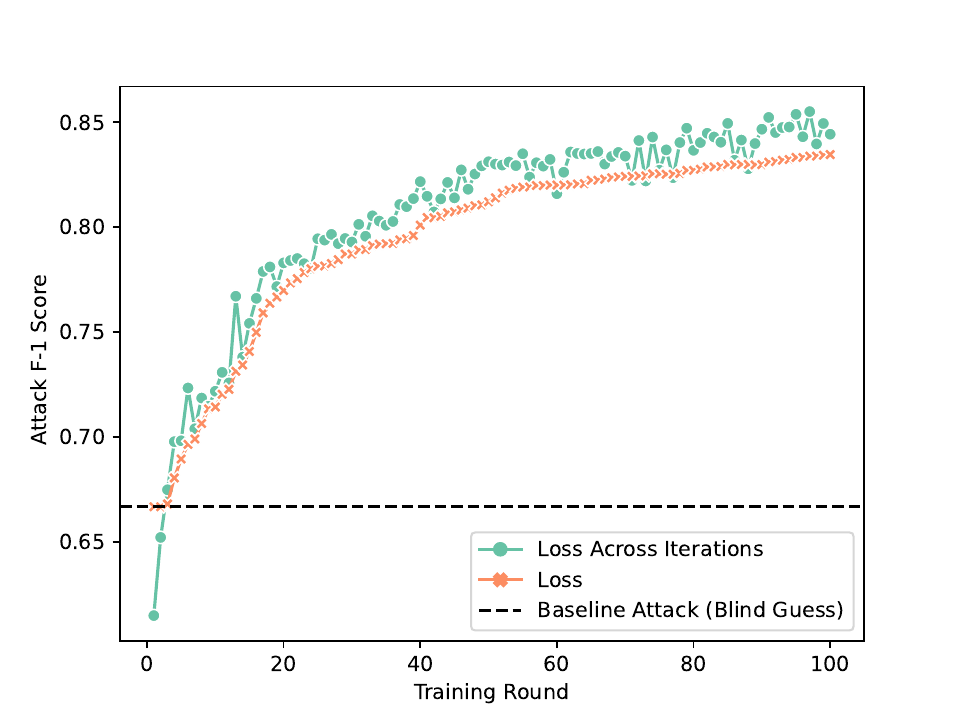}
\caption{Inference risk for the proposed attacks in FL.}
\label{fig:attacks_comparison}
\end{figure}

We begin by comparing the efficacy of our two attacks in the FL setting. Figure~\ref{fig:attacks_comparison} shows results for the 
two proposed variants of the distribution-based attack in the FL setting. Intuitively, the Loss-Across-Rounds Attack should be at least as powerful as the Loss-Threshold Attack, since the former has additional information about the model across its training. Indeed, the Loss-Across-Rounds Attack outperforms the Loss-Threshold Attack nearly across all of the training rounds. However, the gap in performance is negligible for most cases. Similar trends hold in the item-based variants for FL and in the standard training experiments, as shown in the Appendix (Figure~\ref{fig:both_attacks}). Given their similar performance, and the weaker assumptions made by the Loss-Threshold Attack (access to only the final trained model), we default to the latter for evaluations in the rest of the section.

\subsubsection{Differential Privacy}
\label{sec:dp_effect}

One of the most commonly prescribed method for defending against membership inference attack is training ML models with Differential Privacy (DP)~\cite{dwork2006calibrating}. In particular, Federated Learning models can be trained with Local Differential Privacy\cite{warner1965randomized, evfimievski2003limiting,kasiviswanathan08} at various granularities as described before~\cite{abadi16,mcmahan18,liu2020learning,marathesubject}. Algorithms can either provide guarantees at the level of records (Item DP), federation users (User DP), or data subjects (Subject DP). Item DP is implemented using a federated variant of the DP-SGD algorithm~\cite{abadi16}. User DP is another variant of federated DP-SGD that provides user-level local DP~\cite{marathesubject}. Subject DP is the Hierarchical Gradient Averaging algorithm that guarantees subject-level DP~\cite{marathesubject}. HiGradAvgDP builds on the DP-SGD algorithm by Abadi et al~\cite{abadi16}. To obfuscate the contribution of a subject to mini-batch gradients HiGradAvgDP scales down each subject’s mini-batch gradient contribution by averaging it and then clipping that average to the threshold C. This bounds the sensitivity of the algorithm. Gaussian noise is then added at the scale of the clipping threshold.
We evaluate all these algorithms against models trained in FL without any DP, and report results for privacy parameters $\varepsilon=4.0,\delta=10^{-5}$ in Table~\ref{tab:dp_fmnist_results}.

\begin{figure}[ht]
\centering
\begin{subfigure}[b]{\smallfigscale}
    \centering
    \includegraphics[trim={0.2cm 0.1cm 1.6cm 1.3cm},clip,width=\textwidth]{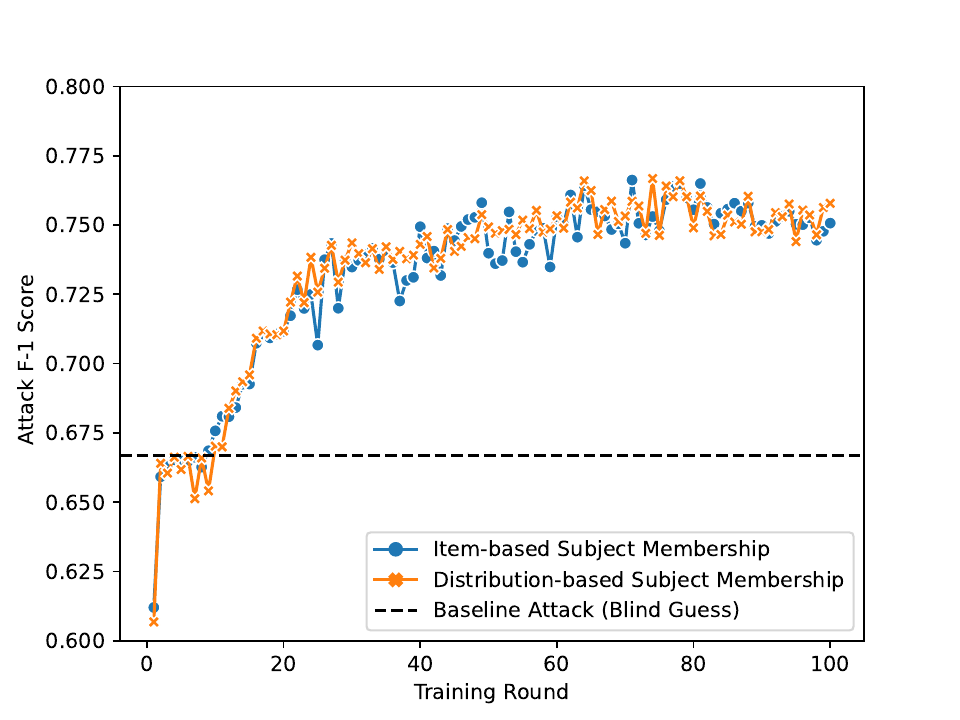}
    \caption{Item-level DP}
    \label{fig:mi_vs_si_fl_item}
\end{subfigure}
\hfill
\begin{subfigure}[b]{\smallfigscale}
    \centering
    \includegraphics[trim={0.2cm 0.1cm 1.6cm 1.3cm},clip,width=\textwidth]{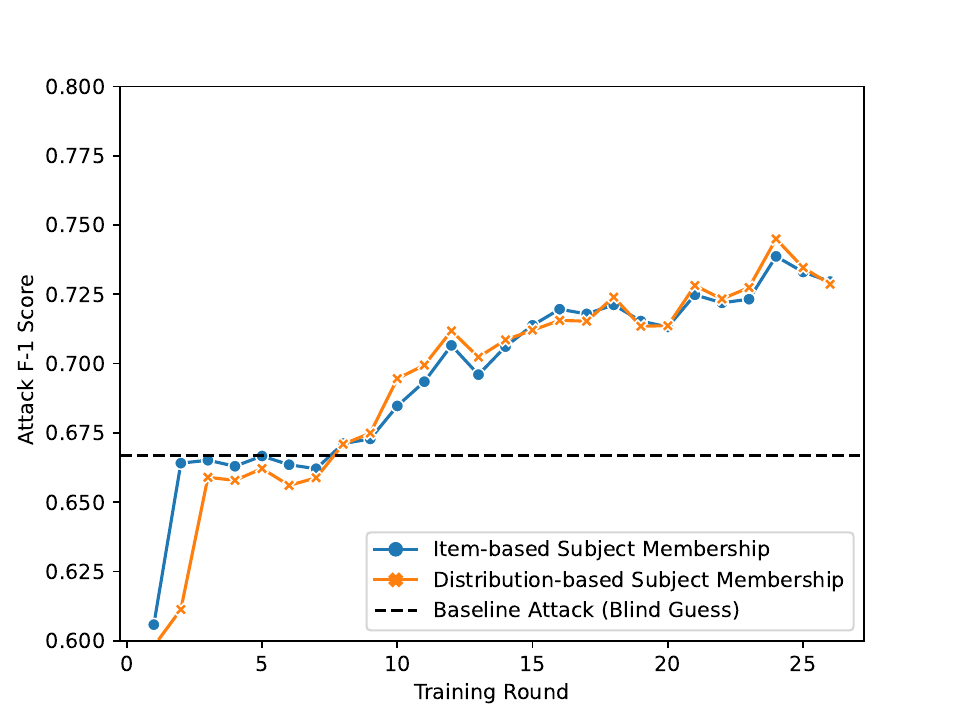}
    \caption{Subject-level DP}
    \label{fig:mi_vs_si_fl_sub}
\end{subfigure}
\caption{Attack F-1 Score across training rounds inferring subject membership using our attack (orange) and via membership inference attacks using exact records (blue), for Item-level DP (\protect{\subref{fig:mi_vs_si_fl_item}}) and Subject-level DP (\protect{\subref{fig:mi_vs_si_fl_sub}}). Both variants of the attacks are equivalent in potency, irrespective of the granularity of DP used for protection.}
\label{fig:mi_vs_si_dp}
\end{figure}

\shortsection{DP Granularities}
User-level \emph{local} DP provides the best protection and completely eliminates inference risk~\cite{marathesubject}, but at the cost of massive drops in task accuracy, rendering it impractical. This protection is expected, since user-level local DP is a strictly stronger notion of privacy than subject-level DP. Subject-level DP, designed exactly for our threat model, lowers inference risk to near-random with a considerable drop in task performance. Item-level DP, as expected, provides the least protection against our distribution-based adversaries.
\begin{table}[h]
    \centering
    \begin{tabular}{lc c c cc}
         \toprule
         \bf \multirow{2}{*}{Granularity} & \multicolumn{4}{c}{\bf Attack} & \bf Task \\
         & Accuracy & Precision & Recall &  F$_1$ & Accuracy \\
         \midrule
         FL&.82&.79&.89&.83&$91.9\pm1.0$\\
         Item DP&.73&.69&.83&.76& $85.1\pm1.5$\\
         User DP&.51&.51&.98&.67& $41.0\pm1.5$\\
         Subject DP&.65&.61&.88&.72& $81.5\pm1.7$\\
        \bottomrule
    \end{tabular}
    \captionsetup{justification=centering}
    \caption{Attack metrics and model task accuracy for vanilla FL and under different DP granularities at privacy budget $\epsilon=4.0$, while using the \textit{Distribution-based Loss-Threshold Attack} on FEMNIST. F$_1$ scores across training rounds are given in the Appendix (Figure~\ref{fig:dp_training_logs}).}
    \label{tab:dp_fmnist_results}
\end{table}
Closer inspection of inference risk under these different granularities of privacy reveals how distribution-based adversaries are just as powerful as Item-based adversaries, even in the presence of these defense mechanisms. (Figure~\ref{fig:mi_vs_si_dp}).

\shortsection{Varying Privacy Budget}
The previous experiment shows that for a privacy budget of $\epsilon=4.0$, the proposed inference attacks maintain residual risk. We next study if further reduction in the privacy budget successfully eliminates this risk. Note that we do not perform this experiment with user-level DP, since in that case, the attack F1 is already close to random at $\epsilon=4.0$.

\begin{figure}[ht]
\centering
\begin{subfigure}[b]{\smallfigscale}
    \centering
    \includegraphics[trim={0.2cm 0.1cm 1.6cm 1.3cm},clip,width=\textwidth]{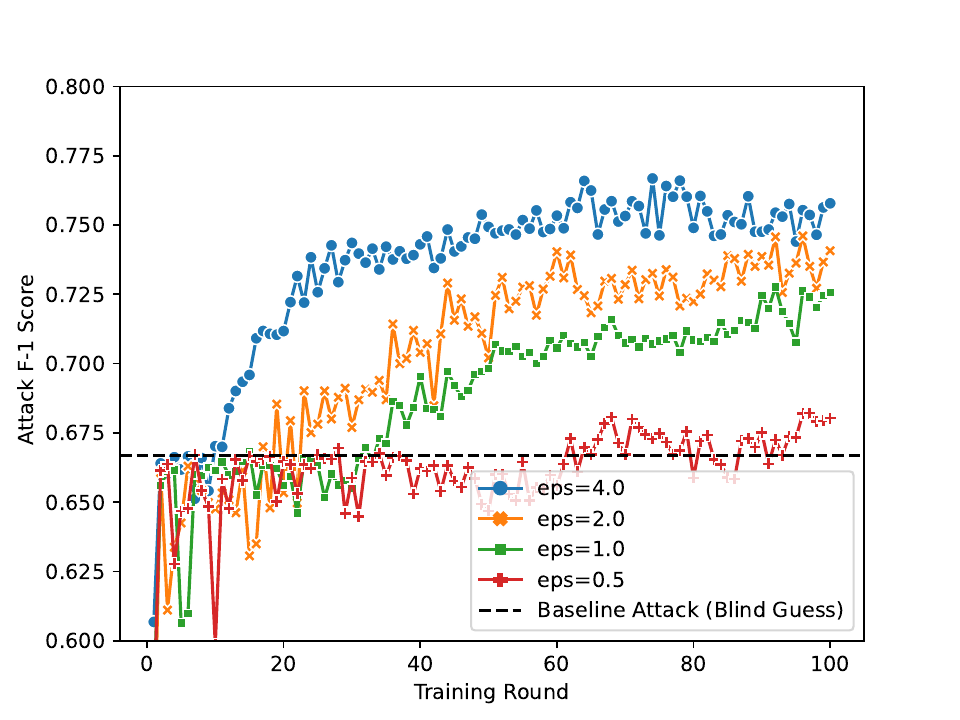}
    \caption{Item-level DP}
    \label{fig:fl_item_vary_eps}
\end{subfigure}
\hfill
\begin{subfigure}[b]{\smallfigscale}
    \centering
    \includegraphics[trim={0.2cm 0.1cm 1.6cm 1.3cm},clip,width=\textwidth]{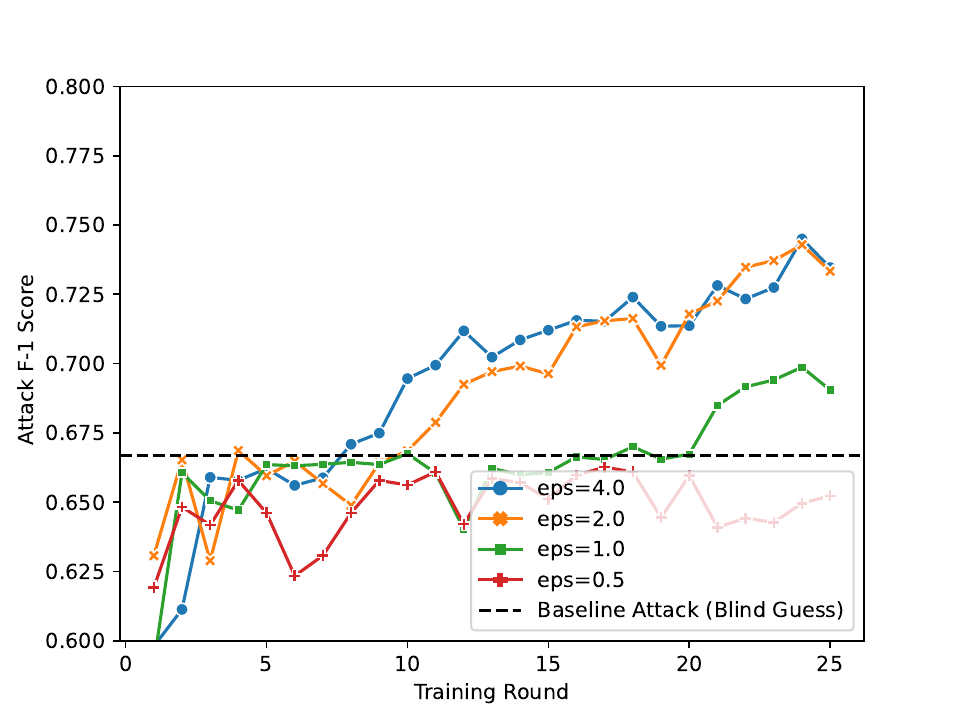}
    \caption{Subject-level DP}
    \label{fig:fl_sub_vary_eps}
\end{subfigure}
\caption{Attack F-1 Score across training rounds inferring subject membership using our attack, for Item-level DP (\protect{\subref{fig:fl_item_vary_eps}}) and Subject-level DP (\protect{\subref{fig:fl_sub_vary_eps}}) with varying levels of protection ($\epsilon$). Both variants of the attacks are equivalent in potency, irrespective of the granularity of DP used for protection. Results for Item-based Subject Membership are provided in the Appendix (Figure~\ref{fig:dp_epsilons_item}).}
\label{fig:dp_epsilons}
\end{figure}

\begin{table}[h]
    \centering
    \begin{tabular}{lc c c c}
         \toprule
         \bf Granularity & $\epsilon=4.0$ & $\epsilon=2.0$ & $\epsilon=1.0$ & $\epsilon=0.5$\\
         \midrule
         Item-level & $85.0\pm1.4$ & $81.9\pm1.6$ & $76.9\pm1.7$ & $68.9\pm2.2$ \\
         Subject-level & $81.5\pm1.7$ & $76.3\pm1.9$ & $69.3\pm2.2$ & $57.4\pm2.4$ \\
         \bottomrule
    \end{tabular}
    \captionsetup{justification=centering}
    \caption{Model task accuracy for different $\epsilon$ values under two DP granularities, while using the \textit{Distribution-based Loss-Threshold Attack} on FEMNIST.}
    \label{tab:dp_fmnist_results_epsilons}
\end{table}
Figure \ref{fig:dp_epsilons} shows that decreasing privacy budget indeed helps protect against these attacks. However, as can be seen from Table \ref{tab:dp_fmnist_results_epsilons}, this added protection comes at the cost of loss in task accuracy. 

\subsection{Shakespeare Experimental Setup} 
\label{sec:exp_setup_shakespeare}

The second dataset used in our evaluation is Shakespeare~\cite{caldas18}, a next-letter prediction task on a corpus of data from classic William Shakespeare plays. The dataset is divided by dialogues of Shakespeare play characters, where each character serves as a federation user. In our evaluation, we treat these play characters as data subjects instead of federation users, and uniformly scatter each subject's data items among all federation users. With a total of $660$ subjects, we split the subjects into member and non-member sets of $330$ data subjects each. Each subject's data is scattered uniformly among $16$ federation users, with no data item assigned to more than one federation user.

\shortsection{Target Model Training} We use a stacked LSTM model with two linear layers at the end for the Shakespeare dataset. Like the FEMNIST experiments, we use half of the data points belonging to each of the 330 member subjects for training, and reserve the remaining half as in-distribution data points to sample from for carrying out the distribution-based attack. 
We train each model for 200 rounds using the Adam~\cite{kingma2014adam} optimizer, with a learning rate of 0.01 and batch size 100.

\subsection{Results on Shakespeare}
\begin{figure*}[ht]
\centering
\begin{subfigure}[b]{0.24\textwidth}
    \centering
    \includegraphics[trim={0.6cm 0.5cm 1.6cm 0.9cm},clip,width=\textwidth]{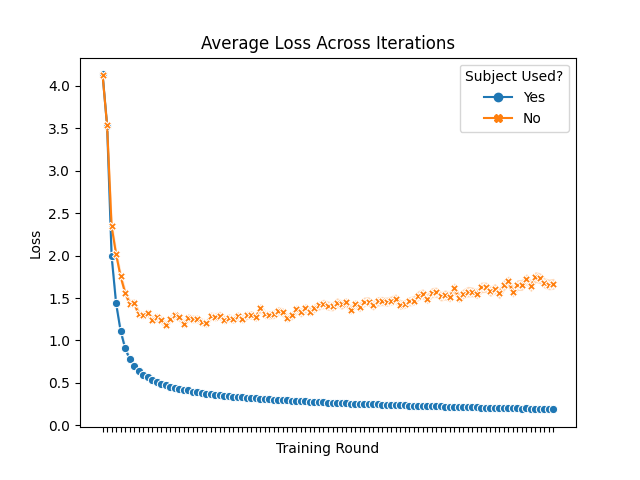}
    \caption{FEMNIST Item-based}
    \label{fig:femnist_item_loss}
\end{subfigure}
\hfill
\begin{subfigure}[b]{0.24\textwidth}
    \centering
    \includegraphics[trim={0.6cm 0.5cm 1.6cm 0.9cm},clip,width=\textwidth]{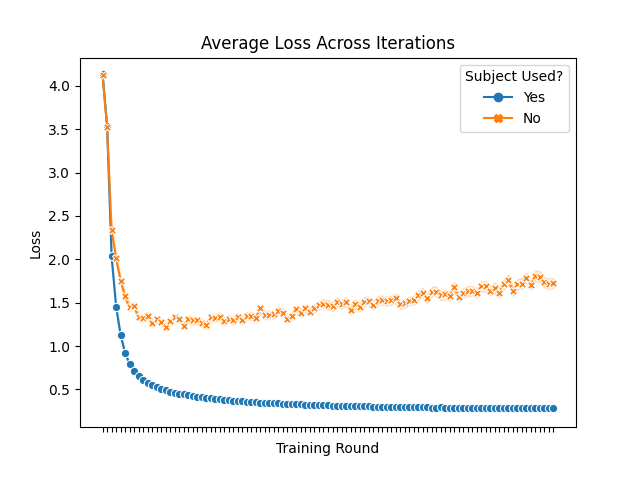}
    \caption{FEMNIST (Distribution)}
    \label{fig:femnist_distribution_loss}
\end{subfigure}
\hfill
\begin{subfigure}[b]{0.24\textwidth}
    \centering
    \includegraphics[trim={0.6cm 0.5cm 1.6cm 0.9cm},clip,width=\textwidth]{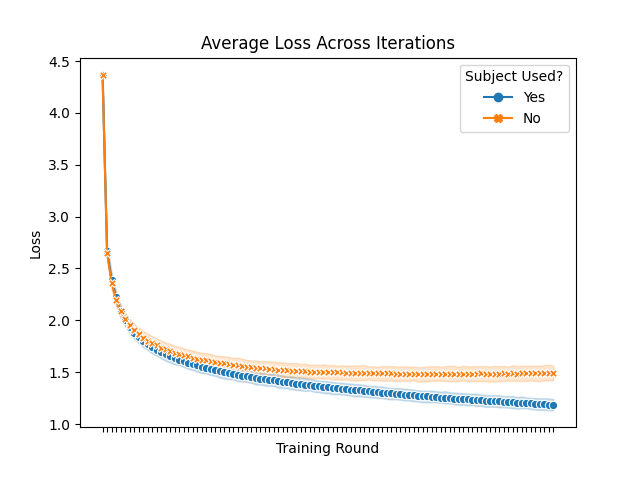}
    \caption{Shakespeare (Item)}
    \label{fig:shakespeare_item_loss}
\end{subfigure}
\hfill
\begin{subfigure}[b]{0.24\textwidth}
    \centering
    \includegraphics[trim={0.6cm 0.5cm 1.6cm 0.9cm},clip,width=\textwidth]{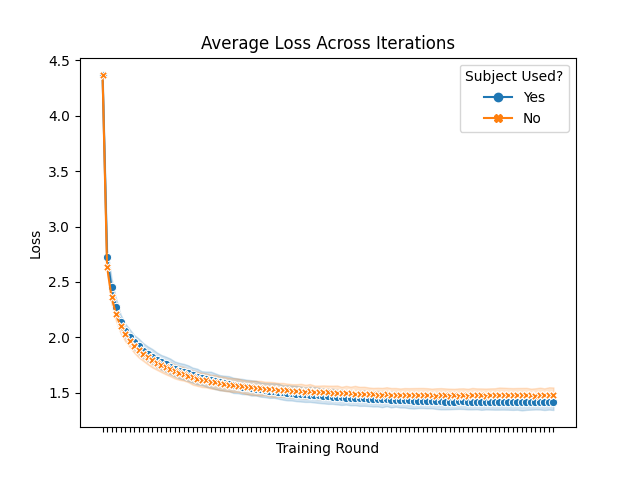}
    \caption{Shakespeare (Distribution)}
    \label{fig:shakespeare_distribution_loss}
\end{subfigure}
\hfill
\caption{Loss over training rounds for the two types of attacks. The difference in loss for the set of subjects that were part of the training data and the ones not included is fairly distinct in case of FEMNIST, whereas for Shakespeare, these two sets are indistinguishable.}
\label{fig:shakespeare_loss}
\end{figure*}

We evaluate the efficacy of our proposed attacks in the federated setting. The \emph{Item-based} version achieves an attack accuracy of $0.51$ while the \emph{Distribution-based} version achieves an attack accuracy of $0.5$. On investigating this attack ineffectiveness on this dataset, we find that the loss values for subjects used in training and the ones not used during the training are almost identical (Figure~\ref{fig:shakespeare_distribution_loss}). One of the reasons this might be happening is because the Shakespeare task is designed to predict the next character (as opposed to a word), and different subjects presumably have a very similar distribution across how they use characters of the English alphabet. This observation is in contrast to FEMNIST, for which we see a clear distinction in loss values across the subjects used in training vs not used for training (Figure~\ref{fig:femnist_distribution_loss}). Since attack accuracy is not high, we do not show results on training with differential privacy on this dataset. 

\section{Synthetic Data}
\label{sec:experiments}

Experiments suggest high subject membership inference leakage in FL, but obtaining real-world, commercial datasets with a clear notion of ``subjects" is non-trivial. It is even more difficult to control federation and data attributes, that can significantly influence subject membership inference risks.  
Although existing synthetic data generators do exist for the FL setting~\cite{caldas2018leaf}, they do not allow fine-grained control over subject-level data generation and its distribution across users. We thus begin with designing our own synthetic data generator for FL (Section~\ref{sec:synthetic_data}). Using FL environments synthesized by our data generator, we evaluate inference risk while varying various aspects of the federation (Section~\ref{sec:results}).

\subsection{Synthetic Federation Data Generator}
\label{sec:synthetic_data}

\begin{figure}
    \centering
    \includegraphics[trim={0 0.5cm 0 0.2cm}, clip,width=0.4\textwidth]{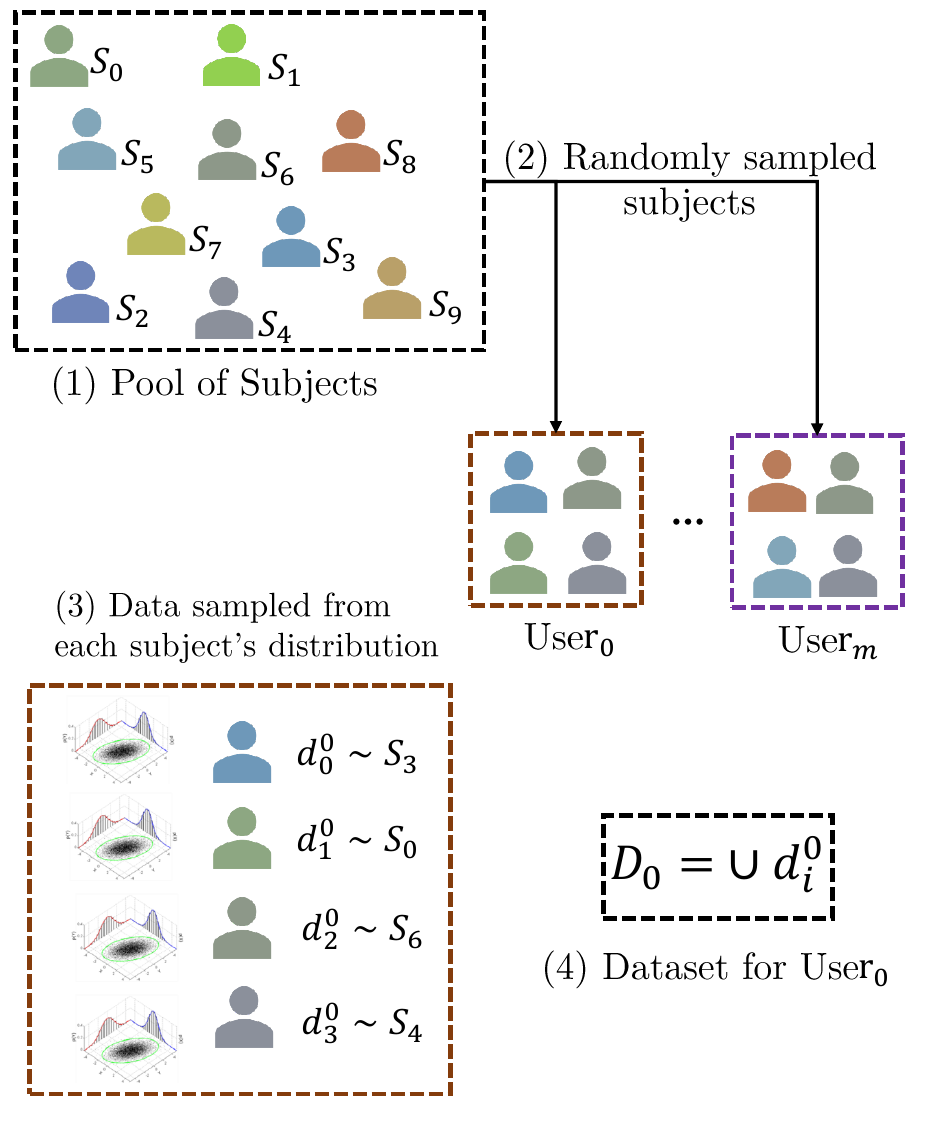}
    \caption{Dataset creation process for our Synthetic Dataset. Each user is assigned subjects at random, and data from each subject's distribution is sampled to generate a user's dataset.}
    \label{fig:dataset_create}
\end{figure}

An ideal configuration setup should allow control over all parameters, even the ones usually fixed for a given dataset (e.g. number of subjects per user, items per subject, items per user).
For a fully controlled federation environment, we design a synthetic dataset generator with multiple controllable parameters, quantifying certain aspects of interest in a federation and study their impact on subject membership inference risk. This generator simulates an entire federation with the given configurable parameters.

We start with a certain controllable dimensionality for the feature space of data. The ground truth label for each data point is computed as the XOR of the features across all dimensions. The idea is to split the grid into a checkboard-like layout, leading to scaling of complexity for the model as we increase dimensions. For a particular data point $x$ with $n$ dimensions:
\begin{align}
    y = \bigoplus_i \mathbb{I}[x_i \geq 0]
\end{align}
The data generation process
(Figure~\ref{fig:dataset_create})
is described below:
\begin{itemize}
    \item [(1)] We model each subject as a parameterized distribution, using a multivariate Gaussian. We generate random (and valid) mean and covariance matrices for each subject, such that no two subjects have the same parameters to their distributions. Additionally, we enforce (achieved by iterative random sampling of subject means until separation requirements are met.) a minimum pair-wise ($L_2 > 0.35$) separation between all of the subject distributions' means to avoid overlap. This separation is set such that it is not too high to make the subjects too distinct and the inference task trivial, yet low enough to be able to tell any two distributions apart.
    \item [(2)] Each user in the federation is then assigned a random sample of subjects. These subjects are sampled from the pool of all subjects with replacement, and thus users can have an overlap in the subjects assigned to them.
    \item [(3)] To construct the user's dataset, data is randomly sampled from distributions of each of the subjects assigned to that particular user. There are two possible extremes when modeling distributions for subjects: each sample being virtually unique and the other with scope for multiple repetitions. The former is more like a patient's blood report readings, while the latter is closer to a customer's shopping cart. We allow for two sampling schemes to capture these two extremes: standard sampling for a multivariate Gaussian and sampling from a Dirichlet process with a multivariate Gaussian as the base distribution, and $\alpha=1$ (hereafter, Dirichlet sampling).
    \item[(4)] The data sampled from each of the user's assigned subjects is then concatenated to form the user's dataset. We repeat this process for all users in the federation.
\end{itemize}
The number of users, total available subjects, number of subjects per user, and data samples per user, are all controllable parameters of our environment.

\subsection{Results on Synthetic Data}

One of the primary motives of this research is to study the impact of different configuration parameters on inference risk. Thus, we choose an extremely strong adversary, with a dataset of a large number of subjects that it knows did and did not participate in the federation. Our results then help us study this empirical upper bound on leakage from the given model(s), even if the adversary somehow computed its threshold(s) using ground truth.

In most of our experiments, we assume the attacker has a wide range of subjects with inclusion/exclusion labels to tune the attack thresholds, and that the included subjects span multiple federation users. This scenario is plausible when the adversary is the federation server (which is what we assume for the rest of the paper, unless specified otherwise), as some subjects used in the federation are likely already known from side channels. If the adversary is not the federation server, a dataset spanning multiple federation users may still exist, or the adversary can begin with an educated guess of membership labels.

The success of inference attacks can depend on several factors:
\begin{itemize}
    \item \textbf{Data Properties:} dimensionality and sampling distribution
    \item \textbf{Model Design and Training:} model architecture and number of training rounds
    \item \textbf{Federation Properties:} number of users, subjects, and datapoints
\end{itemize}
For a comprehensive evaluation of how these factors influence subject membership inference risk, we generate 720 configurations by varying all of the above parameters systematically on the synthetic dataset. The exact configuration values are given in Table~\ref{table:synthetic_configs}. This extensive grid search is a one-of-its-kind study for Federated Learning systems and is meant to expand our understanding of how certain factors, both in and out of the model trainer's control, can influence privacy leakage. The existing literature on inference attacks is limited to using a handful of datasets, which does not enable a new kind of scientific exploration that we undertake. 

\begin{table}[h]
\centering
\begin{tabular}{l r} 
 \toprule
 Configurable & Values Experimented\\
 \midrule
 Sampling Mechanism & \{Normal, Dirichlet\} \\
 Data Dimensionality & \{2, 50, 250, 1000\} \\
 \midrule
 Model: Number of Layers & \{1, 2, 3\}\\
 Model: Number of Epochs & [1, 50]\\
 \midrule
 Users & \{10, 100\} \\
 Subjects per User & \{10, 100, 500\} \\
 Items per User & \{500, 2000, 10000\} \\
 \bottomrule
\end{tabular}
\caption{Variables for the Synthetic Dataset that we experiment with. Each of these are tried simultaneously, thus yielding all 720 possible configurations with these values.}
\label{table:synthetic_configs}
\end{table}

\subsubsection{Attack Success and High Risk Configurations}
\label{sec:exploratory_results}

We run both the attacks described in Section~\ref{sec:method} on these configurations. The $F_1$ attack scores are no better than random guessing for half of the configurations, while for the other half they go as high as $\sim0.95$. The appendix shows results on a few example configurations that represent a good variety in the various environmental variables like sampling mechanism, data dimensionality, and model capacity. We note that the privacy leakage risk for most real world ML systems fall somewhere on this spectrum. 

To better understand what combinations of the various parameters may make the overall federation more susceptible to these inference attacks, we choose to look at highly successful attacks: ones with both precision and $F_1$ scores $>0.9$. Close analysis of the filtered configurations yields some common attributes:
\begin{itemize}
    \item High data dimensionality: 1000
    \item Dirichlet sampling while generating data
    \item Large model architectures: $\geq 3$ hidden layers, and 
    \item Models trained for many rounds: $\geq 20$.
\end{itemize}
Looking out for these attributes can help machine learning practitioners identify cases that may be highly susceptible to subject membership inference attacks.

\subsubsection{Grid-Search Results}
\label{sec:grid_results}

We now present results on the success of subject membership inference attack aggregated over all 720 synthetic configurations, broken down by different factors. Our early exploratory experiments on a subset of configurations indicate that the three attacks are often correlated. Concretely, for the configurations mentioned in Section~\ref{sec:exploratory_results}, we observe a positive correlation of 
$\sim0.976$ between \textit{Loss-Threshold} and \textit{Loss-Across-Rounds} attacks.
Thus for simplicity and computational efficiency, we only report results for the \textit{Loss-Threshold} Attack.

This grid-based experimental protocol also helps us uncover some important trends, which can be used to provide practical guidelines to ML practitioners about the vulnerability of their FL setup or model architectures (Appendix~\ref{sec:specific_configs_synthetic}).

\subsubsection{Data Properties}

\begin{figure}
    \centering
    \includegraphics[trim={0cm 0.1cm 1.6cm 1.3cm},clip,width=\smallfigscale]{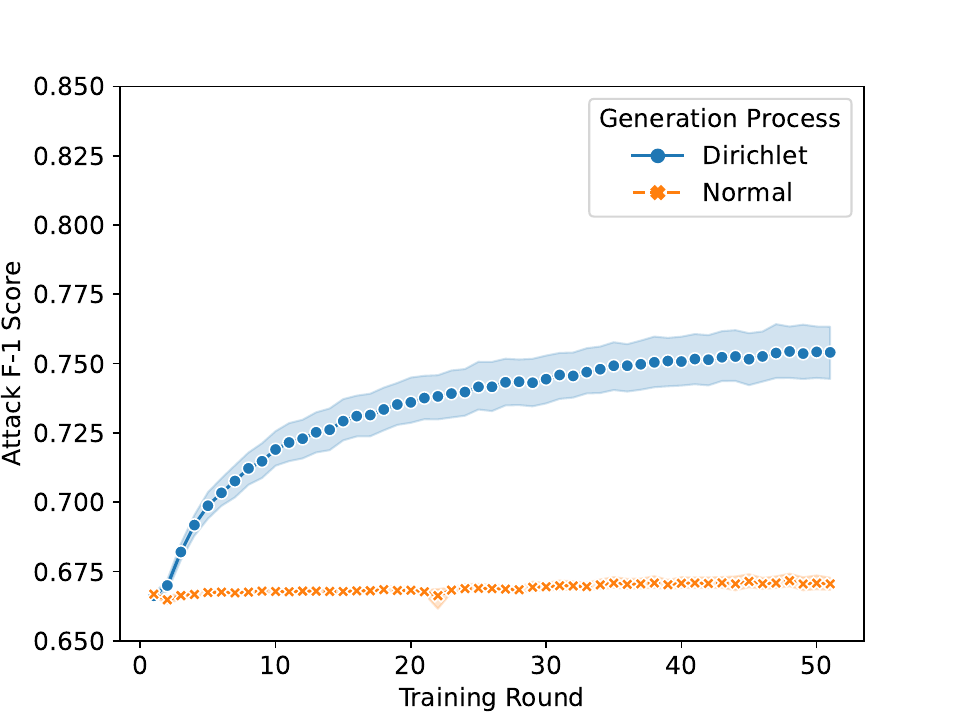}
    \caption{Attack F-1 Score across training rounds for datasets generation with Standard and Dirichlet Sampling. Dirichlet sampling increases susceptibility to subject membership inference significantly, which is not surprising.}
    \label{fig:sampling}
\end{figure}

\shortsection{Sampling Mechanism}
We plot Attack F-1 scores across training rounds, for data distributions with standard and Dirichlet sampling (Figure~\ref{fig:sampling}). We observe Dirichlet sampling to exhibit a significantly higher inference risk than the case of regular sampling. This is expected since repeated samples would make inferring a subject's membership easier, almost reducing it to datapoint membership inference. These sampling mechanisms represent two  extreme cases possible in real-world federation systems: each datum being sampled uniquely (like blood cell counts) versus high density around specific data points (like grocery store purchases). Real-world datasets would be somewhere between these two, and having results for them both gives a good sense of the expected range of inference risk for real-world datasets.

\begin{figure}
    \centering
    \includegraphics[trim={0cm 0.1cm 1.6cm 1.3cm},clip,width=\smallfigscale]{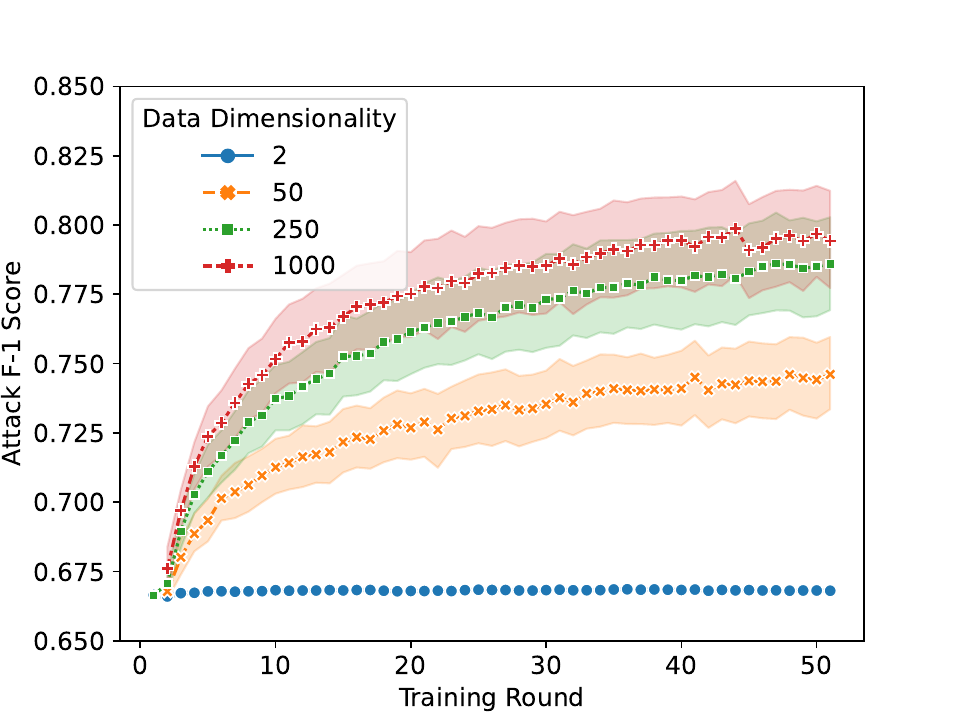}
    \caption{Attack F-1 Score across training rounds for datasets with different feature dimensionality. Larger data dimensionality leads to more sparsity in subject distributions, making it easier to distinguish between them.}
    \label{fig:dimensionality}
\end{figure}

\shortsection{Dimensionality}
Inference risk seems to correlate positively with the dimensionality of the feature space (Figure~\ref{fig:dimensionality}), with stagnation in the F-1 scores for inference as the dimensionality increases beyond a certain point. Subject distributions in lower dimensions are likely to be closer to each other. On the other hand, the same number of distributions in a higher-dimensional space would be distributed much more sparsely, owing to the curse of dimensionality. Thus, the latter would be understandably easier to distinguish than the former. Model trainers thus need to be cautious when working with high dimensional data since that may make them highly susceptible to such inference attacks.

\subsubsection{Model Design and Training}

\begin{figure}
    \centering
    \includegraphics[trim={0cm 0.1cm 1.6cm 1.3cm},clip,width=\smallfigscale]{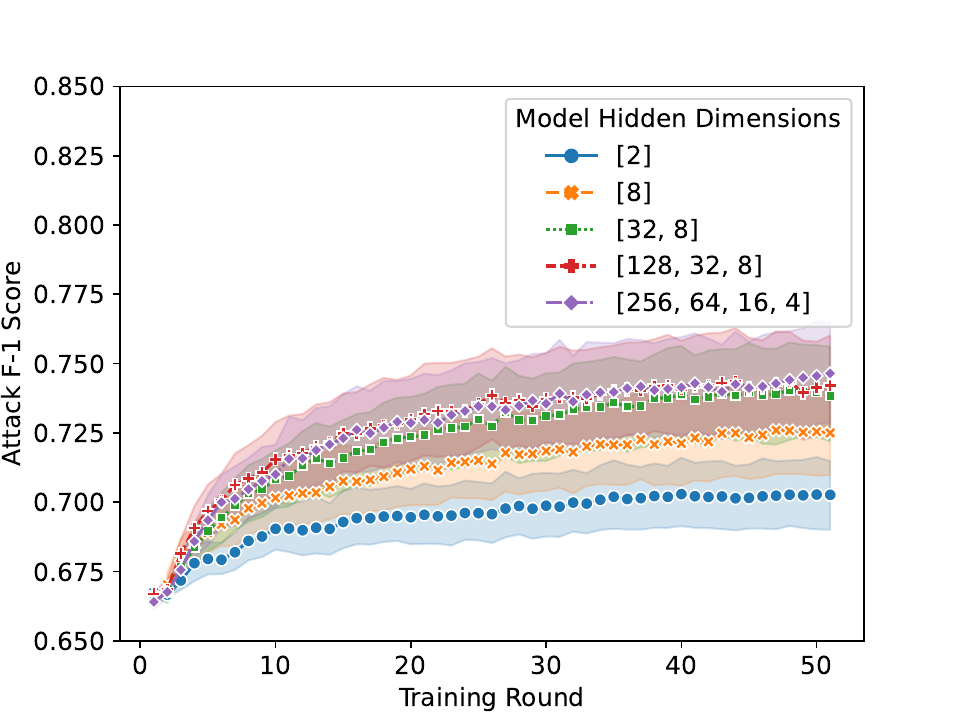}
    \caption{Attack F-1 Score across training rounds for datasets with different model architectures. Hidden dimensions refers to the number and sizes of intermediate layers for the neural networks used. Increasing model capacity may make the model more capable of learning subject distributions, increasing their risk to subject membership inference.}
    \label{fig:hidden_dims}
\end{figure}

\shortsection{Model Complexity}
We vary model complexity by adjusting both the number of layers and neurons per layer, going from a single hidden layer neural network up to one with four hidden layers (Figure~\ref{fig:hidden_dims}). Inference risk seems to increase model complexity but plateaus beyond model complexity required for the task. The risk increases as we increase the number of neurons for the same one-hidden-layer architecture and then again on adding an additional hidden layer. However, more complex models exhibit almost similar inference risk. Interestingly, inference risk for the under-parameterized models is only slightly better than random guessing, suggesting it may be in the model trainers' interest to use models that are not too complex for a given task. This suggestion is in the model trainer's own interest, since using a smaller model also makes the model less likely to overfit.

\shortsection{Model Training}
Similar to trends with model complexity, we observe that inference risk increases as the model continue to train and then plateaus towards the latter half of training rounds, which is a few rounds after the model's loss has converged on both train and test data. These observations are clearly visible in all of the previous figures, and especially in Figure~\ref{fig:hidden_dims}. Based on these observations, it would make sense not to train the model for too many rounds- only enough to achieve satisfactory performance. Such a decision may hurt the model trainer, since some studies in the literature~\cite{papyan2020prevalence} demonstrate how training beyond convergence can confer benefits like better robustness, generalization, and interpretability.

\subsubsection{Federation Properties}

\begin{figure}
\centering
\begin{subfigure}[b]{\smallfigscale}
    \centering
    \includegraphics[trim={0cm 0.1cm 1.6cm 1.3cm},clip,width=\textwidth]{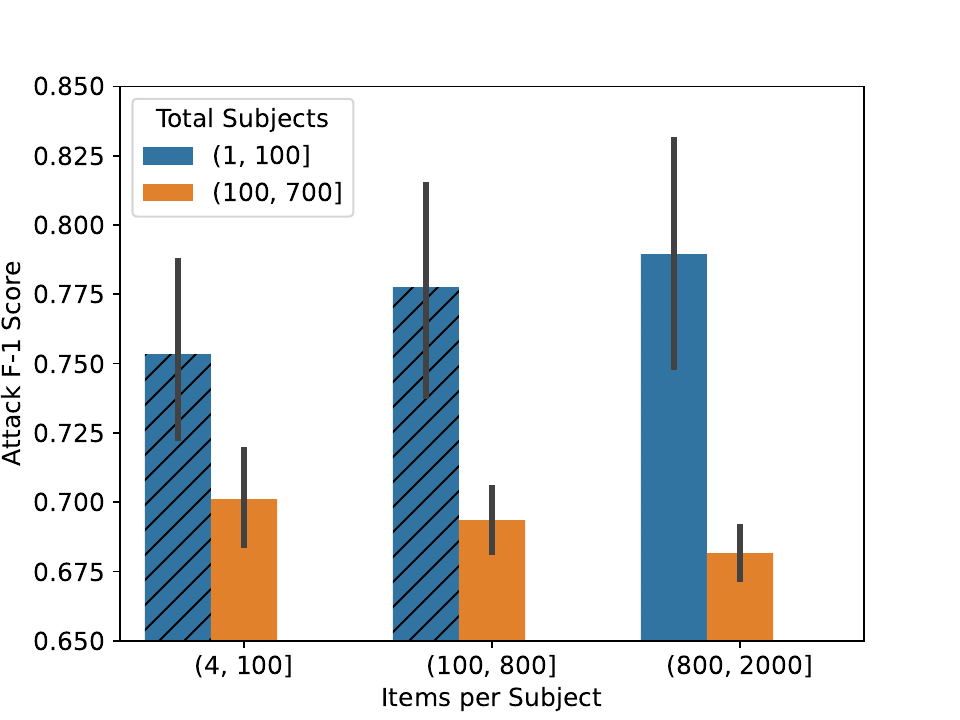}
    \caption{10 Subjects per User}
    \label{fig:fed_su_10}
\end{subfigure}
\hfill
\begin{subfigure}[b]{\smallfigscale}
    \centering
    \includegraphics[trim={0cm 0.1cm 1.6cm 1.3cm},clip,width=\textwidth]{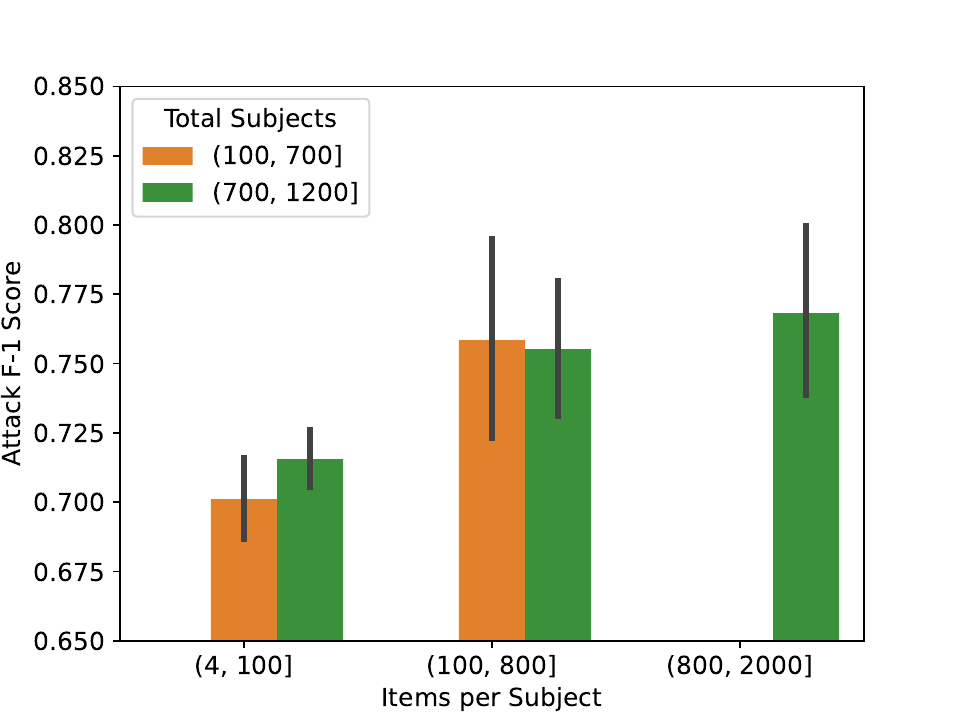}
    \caption{100/500 Subjects per User}
    \label{fig:fed_su_mt10}
\end{subfigure}
\hfill
\caption{Attack F-1 Scores while varying number of total subjects and items per subject, for 10 subjects per user (\protect{\subref{fig:fed_su_10}}) and 100 subjects per user (\protect{\subref{fig:fed_su_mt10}}) in the Federation. Properties of the Federation have a complicated effect on inference risk, which can be decrypted by binning results according to the total number of subjects and analyzing.}
\label{fig:fed_properties}
\end{figure}

For a given number of data points corresponding to a subject, the underlying federation can have several different configurations: different number of users, subjects per user, as well as items per user. Although none of these are controlled by an adversary, understanding how they impact subject membership inference risk can be advantageous in both designing and understanding such attacks. We study these trends across varying parameters of the configuration setup and observe very peculiar trends. We split our analyses into two categories: \textit{Few Subjects per User} (10) and \textit{Many Subjects per User} (100/500). We further calculate the total number of items per subject in each configuration, and bin them into three categories: $(4, 100]$ (\textit{low}), $(100, 800]$ (\textit{medium}), and $(800, 2000]$ (\textit{high}).

\shortsection{Few Subjects per User}
For the case with only a few subjects per user (Figure~\ref{fig:fed_su_10}), inference risk (Y-axis) is higher for cases with fewer total subjects (blue), compared to settings with more total subjects in the federation (orange). This trend is expected, as having more (subject) distributions in the same feature co-domain would make overlap between distributions more likely, making it harder for an adversary to distinguish between any two distributions. Attack F-1 scores change as the number of items per subject (X-axis) increase, but the trends are somewhat conflicting for the \textit{low} and \textit{medium} cases of items per subject. For the former, the F-1 scores increase with increase in items. This is expected, since having more items per subject would make it more likely for the model to generalize well to a given subject's distribution (as opposed to overfitting to a few points), making it easier for an adversary to infer membership. Attack scores decrease for the \textit{medium} case of items per subject, but within error of margin,

\shortsection{Many Subjects per User}
When we have a sufficiently high number of subjects per user, we observe an increase in risk as we increase the number of items per subject (Figure~\ref{fig:fed_su_mt10}). The gains in attack performance too taper off once there are sufficiently large number of items per subject (\textit{medium} v/s \textit{high}) Since the total number of subjects in the system is high enough, the effects of potential overlap between subject distributions (mentioned earlier) start to converge; the two cases (orange and green) are thus not affected much by an increase in the total number of subjects and are close in their performance.

Our analyses show how configurations with a lot of subjects in the federation increase in susceptibility to subject membership inference as the data available per subject increases. At the same time, configurations with few subjects in the federation are highly likely to leak subject membership.

\section{Conclusion}
\label{sec:conclusion}

Privacy in Federated Learning is typically only studied for individual data items or users participating in the federation. However, in complex cross-silo FL settings, we ultimately care about protecting the privacy of individual data subjects, whose vulnerability to privacy attacks increases when the organizations they interact with choose to form federations. Ours is the first work to propose \emph{subject-level} membership inference attacks, which can aid in the empirical measurement of data subject's privacy leakage.

We show that even under much weaker assumptions about the adversary, it is possible to retrieve subject membership information from a wide variety of Federated Learning models. Instead of having access to exact potential records from training, the ability to draw samples from the subject's data distribution, and knowledge of a handful of subjects' membership is sufficient to execute strong subject membership inference attacks. We show that mitigating these attacks using Differential Privacy is challenging and comes at a heavy cost to task accuracy.

Using our first-of-its-kind, synthetic data generator-based study,
we vary three main aspects of the system across several hundred FL configurations---data, model, and FL structure, and find that factors like data distribution and dimensionality, model complexity, training protocols, the size and composition of the federation in terms of the number of users, data subjects and data items, all have a substantial impact on the attack accuracy.
This study provides invaluable practical guidance to model designers and ML practitioners on what makes their models more vulnerable. 
 
Attacks that have access to the final state of a trained model have access to less information than ones with access to each training round. The fact that our evaluations yield similar performance for attacks in these two settings implies the existence of stronger attacks that can better harness information across rounds. Similarly, we would like to explore attacks that assume complete white box access to the models, which is also common in many cross-silo FL settings.
We hope that this study on subject membership inference will help ML practitioners focus on and protect the most important asset in the FL data ecosystem - the people.

\bibliographystyle{IEEEtran}
\bibliography{references}

\begin{thebibliography}{10}
\providecommand{\url}[1]{#1}
\csname url@samestyle\endcsname
\providecommand{\newblock}{\relax}
\providecommand{\bibinfo}[2]{#2}
\providecommand{\BIBentrySTDinterwordspacing}{\spaceskip=0pt\relax}
\providecommand{\BIBentryALTinterwordstretchfactor}{4}
\providecommand{\BIBentryALTinterwordspacing}{\spaceskip=\fontdimen2\font plus
\BIBentryALTinterwordstretchfactor\fontdimen3\font minus
  \fontdimen4\font\relax}
\providecommand{\BIBforeignlanguage}[2]{{%
\expandafter\ifx\csname l@#1\endcsname\relax
\typeout{** WARNING: IEEEtran.bst: No hyphenation pattern has been}%
\typeout{** loaded for the language `#1'. Using the pattern for}%
\typeout{** the default language instead.}%
\else
\language=\csname l@#1\endcsname
\fi
#2}}
\providecommand{\BIBdecl}{\relax}
\BIBdecl

\bibitem{shokri2017membership}
R.~Shokri, M.~Stronati, C.~Song, and V.~Shmatikov, ``{Membership Inference
  Attacks Against Machine Learning Models},'' in \emph{2017 IEEE symposium on
  security and privacy (SP)}.\hskip 1em plus 0.5em minus 0.4em\relax IEEE,
  2017, pp. 3--18.

\bibitem{dwork06a}
C.~Dwork, ``Differential privacy,'' in \emph{Automata, Languages and
  Programming, 33rd International Colloquium, {ICALP}}, 2006, pp. 1--12.

\bibitem{yeom2018privacy}
S.~Yeom, I.~Giacomelli, M.~Fredrikson, and S.~Jha, ``{Privacy Risk in Machine
  Learning: Analyzing the Connection to Overfitting},'' in \emph{2018 IEEE 31st
  Computer Security Foundations symposium (CSF)}.\hskip 1em plus 0.5em minus
  0.4em\relax IEEE, 2018, pp. 268--282.

\bibitem{jayaraman2021revisiting}
B.~Jayaraman, L.~Wang, K.~Knipmeyer, Q.~Gu, and D.~Evans, ``{Revisiting
  Membership Inference Under Realistic Assumptions},'' \emph{Proceedings on
  Privacy Enhancing Technologies}, vol. 2021, no.~2, 2021.

\bibitem{abowd20222020}
J.~M. Abowd, R.~Ashmead, R.~Cumings-Menon, S.~Garfinkel, M.~Heineck, C.~Heiss,
  R.~Johns, D.~Kifer, P.~Leclerc, A.~Machanavajjhala \emph{et~al.}, ``{The 2020
  Census Disclosure Avoidance System TopDown Algorithm},'' \emph{arXiv preprint
  arXiv:2204.08986}, 2022.

\bibitem{thakurta2017learning}
A.~G. Thakurta, A.~H. Vyrros, U.~S. Vaishampayan, G.~Kapoor, J.~Freudiger,
  V.~R. Sridhar, and D.~Davidson, ``Learning new words,'' \emph{Granted US
  Patents}, vol. 9594741, 2017.

\bibitem{chen2023face}
M.~Chen, Z.~Zhang, T.~Wang, M.~Backes, and Y.~Zhang, ``Face-auditor: Data
  auditing in facial recognition systems,'' in \emph{USENIX Security
  Symposium}, 2023.

\bibitem{marathesubject}
\BIBentryALTinterwordspacing
V.~J. Marathe, P.~Kanani, and D.~W. Peterson, ``Subject level differential
  privacy with hierarchical gradient averaging,'' in \emph{International
  Workshop on Federated Learning: Recent Advances and New Challenges in
  Conjunction with NeurIPS 2022}, 2022. [Online]. Available:
  \url{https://openreview.net/forum?id=vvQGXlHKkIG}
\BIBentrySTDinterwordspacing

\bibitem{suri2022formalizing}
A.~Suri and D.~Evans, ``{Formalizing and Estimating Distribution Inference
  Risks},'' in \emph{Privacy Enhancing Technologies Symposium}, 2022.

\bibitem{mcmahan2017communication}
B.~McMahan, E.~Moore, D.~Ramage, S.~Hampson, and B.~A. y~Arcas,
  ``{Communication-Efficient Learning of Deep Networks from Decentralized
  Data},'' in \emph{Artificial intelligence and statistics}.\hskip 1em plus
  0.5em minus 0.4em\relax PMLR, 2017, pp. 1273--1282.

\bibitem{kairouz2021advances}
P.~Kairouz, H.~B. McMahan, B.~Avent, A.~Bellet, M.~Bennis, A.~N. Bhagoji,
  K.~Bonawitz, Z.~Charles, G.~Cormode, R.~Cummings \emph{et~al.}, ``Advances
  and open problems in federated learning,'' \emph{Foundations and
  Trends{\textregistered} in Machine Learning}, vol.~14, no. 1--2, pp. 1--210,
  2021.

\bibitem{abadi16}
M.~Abadi, A.~Chu, I.~Goodfellow, H.~B. McMahan, I.~Mironov, K.~Talwar, and
  L.~Zhang, ``{Deep Learning with Differential Privacy},'' in \emph{Proceedings
  of the 2016 ACM SIGSAC Conference on Computer and Communications Security},
  2016, pp. 308--318.

\bibitem{mcmahan18}
H.~B. McMahan, D.~Ramage, K.~Talwar, and L.~Zhang, ``{Learning Differentially
  Private Recurrent Language Models},'' in \emph{6th International Conference
  on Learning Representations, {ICLR} 2018}, 2018.

\bibitem{liu2020learning}
Y.~Liu, A.~T. Suresh, F.~X.~X. Yu, S.~Kumar, and M.~Riley, ``Learning discrete
  distributions: user vs item-level privacy,'' \emph{Advances in Neural
  Information Processing Systems}, 2020.

\bibitem{caldas2018leaf}
S.~Caldas, S.~M.~K. Duddu, P.~Wu, T.~Li, J.~Kone{\v{c}}n{\`y}, H.~B. McMahan,
  V.~Smith, and A.~Talwalkar, ``{LEAF: A Benchmark for Federated Settings},''
  \emph{arXiv preprint arXiv:1812.01097}, 2018.

\bibitem{dwork2006calibrating}
C.~Dwork, F.~McSherry, K.~Nissim, and A.~Smith, ``{Calibrating Noise to
  Sensitivity in Private Data Analysis},'' in \emph{Theory of cryptography
  conference}.\hskip 1em plus 0.5em minus 0.4em\relax Springer, 2006, pp.
  265--284.

\bibitem{Truong2021-cd}
N.~Truong, K.~Sun, S.~Wang, F.~Guitton, and Y.~Guo, ``Privacy preservation in
  federated learning: An insightful survey from the gdpr perspective,''
  \emph{Computers \& Security}, vol. 110, p. 102402, 2021.

\bibitem{jegorova2022survey}
M.~Jegorova, C.~Kaul, C.~Mayor, A.~Q. O'Neil, A.~Weir, R.~Murray-Smith, and
  S.~A. Tsaftaris, ``Survey: Leakage and privacy at inference time,''
  \emph{IEEE Transactions on Pattern Analysis and Machine Intelligence}, 2022.

\bibitem{liu2022ml}
Y.~Liu, R.~Wen, X.~He, A.~Salem, Z.~Zhang, M.~Backes, E.~De~Cristofaro,
  M.~Fritz, and Y.~Zhang, ``$\{$ML-Doctor$\}$: Holistic risk assessment of
  inference attacks against machine learning models,'' in \emph{31st USENIX
  Security Symposium (USENIX Security 22)}, 2022, pp. 4525--4542.

\bibitem{fredrikson2015model}
M.~Fredrikson, S.~Jha, and T.~Ristenpart, ``{Model Inversion Attacks that
  Exploit Confidence Information and Basic Countermeasures},'' in
  \emph{Proceedings of the 22nd ACM SIGSAC Conference on Computer and
  Communications Security}, 2015, pp. 1322--1333.

\bibitem{zhang2020secret}
Y.~Zhang, R.~Jia, H.~Pei, W.~Wang, B.~Li, and D.~Song, ``{The Secret Revealer:
  Generative Model-Inversion Attacks Against Deep Neural Networks},'' in
  \emph{Proceedings of the IEEE/CVF Conference on Computer Vision and Pattern
  Recognition}, 2020, pp. 253--261.

\bibitem{ateniese2015hacking}
G.~Ateniese, L.~V. Mancini, A.~Spognardi, A.~Villani, D.~Vitali, and G.~Felici,
  ``Hacking {S}mart {M}achines with {S}marter {O}nes: {H}ow to {E}xtract
  {M}eaningful {D}ata from {M}achine {L}earning {C}lassifiers,''
  \emph{International Journal of Security and Networks}, vol.~10, no.~3, pp.
  137--150, 2015.

\bibitem{ganju2018property}
K.~Ganju, Q.~Wang, W.~Yang, C.~A. Gunter, and N.~Borisov, ``{Property Inference
  Attacks on Fully Connected Neural Networks using Permutation Invariant
  Representations},'' in \emph{Proceedings of the 2018 ACM SIGSAC conference on
  computer and communications security}, 2018, pp. 619--633.

\bibitem{Shokri2017-tg}
R.~Shokri, M.~Stronati, C.~Song, and V.~Shmatikov, ``{Membership Inference
  Attacks Against Machine Learning Models},'' 2017.

\bibitem{Salem2019-pk}
A.~Salem, Y.~Zhang, M.~Humbert, P.~Berrang, M.~Fritz, and M.~Backes,
  ``{{ML-Leaks}: Model and Data Independent Membership Inference Attacks and
  Defenses on Machine Learning Models},'' 2019.

\bibitem{Shejwalkar_undated-lk}
V.~Shejwalkar, H.~A. Inan, A.~Houmansadr, and R.~Sim, ``Membership inference
  attacks against {NLP} classification models,'' in \emph{NeurIPS Workshop on
  Privacy in Machine Learning}, 2021.

\bibitem{zhu2019deep}
L.~Zhu, Z.~Liu, and S.~Han, ``{Deep Leakage from Gradients},'' \emph{Advances
  in Neural Information Processing Systems}, vol.~32, 2019.

\bibitem{mo2020layer}
F.~Mo, A.~Borovykh, M.~Malekzadeh, H.~Haddadi, and S.~Demetriou, ``{Layer-wise
  Characterization of Latent Information Leakage in Federated Learning},''
  \emph{arXiv preprint arXiv:2010.08762}, 2020.

\bibitem{Geiping2020-nn}
J.~Geiping, H.~Bauermeister, H.~Dr{\"o}ge, and M.~Moeller, ``{Inverting
  Gradients -- How easy is it to break privacy in federated learning?}''
  \emph{Advances in Neural Information Processing Systems}, 2020.

\bibitem{geng2023improved}
J.~Geng, Y.~Mou, Q.~Li, F.~Li, O.~Beyan, S.~Decker, and C.~Rong, ``Improved
  gradient inversion attacks and defenses in federated learning,'' \emph{IEEE
  Transactions on Big Data}, 2023.

\bibitem{huang2021evaluating}
Y.~Huang, S.~Gupta, Z.~Song, K.~Li, and S.~Arora, ``{Evaluating Gradient
  Inversion Attacks and Defenses in Federated Learning},'' \emph{Advances in
  Neural Information Processing Systems}, vol.~34, 2021.

\bibitem{Wainakh2021-io}
A.~Wainakh, F.~Ventola, T.~M{\"u}{\ss}ig, J.~Keim, C.~G. Cordero, E.~Zimmer,
  T.~Grube, K.~Kersting, and M.~M{\"u}hlh{\"a}user, ``{User Label Leakage from
  Gradients in Federated Learning},'' \emph{arXiv preprint arXiv:2105.09369},
  2021.

\bibitem{Liu2021-qo}
Y.~Liu, X.~Zhu, J.~Wang, and J.~Xiao, ``{A Quantitative Metric for Privacy
  Leakage in Federated Learning},'' in \emph{{ICASSP} 2021 - 2021 {IEEE}
  International Conference on Acoustics, Speech and Signal Processing
  ({ICASSP})}, Jun. 2021, pp. 3065--3069.

\bibitem{Hitaj2017-tq}
B.~Hitaj, G.~Ateniese, and F.~Perez-Cruz, ``{Deep Models Under the {GAN}:
  Information Leakage from Collaborative Deep Learning},'' in \emph{Proceedings
  of the 2017 {ACM} {SIGSAC} Conference on Computer and Communications
  Security}, ser. CCS '17.\hskip 1em plus 0.5em minus 0.4em\relax New York, NY,
  USA: Association for Computing Machinery, Oct. 2017, pp. 603--618.

\bibitem{nguyen2023active}
T.~Nguyen, P.~Lai, K.~Tran, N.~Phan, and M.~T. Thai, ``Active membership
  inference attack under local differential privacy in federated learning,'' in
  \emph{International Conference on Artificial Intelligence and
  Statistics}.\hskip 1em plus 0.5em minus 0.4em\relax PMLR, 2023, pp.
  5714--5730.

\bibitem{Wang2019-qy}
Z.~Wang, M.~Song, Z.~Zhang, Y.~Song, Q.~Wang, and H.~Qi, ``{Beyond Inferring
  Class Representatives: {User-Level} Privacy Leakage From Federated
  Learning},'' in \emph{{IEEE} {INFOCOM} 2019 - {IEEE} Conference on Computer
  Communications}, Apr. 2019, pp. 2512--2520.

\bibitem{mahloujifar2021membership}
S.~Mahloujifar, H.~A. Inan, M.~Chase, E.~Ghosh, and M.~Hasegawa, ``{Membership
  Inference on Word Embedding and Beyond},'' \emph{arXiv preprint
  arXiv:2106.11384}, 2021.

\bibitem{wang2022poisoning}
Z.~Wang, Y.~Huang, M.~Song, L.~Wu, F.~Xue, and K.~Ren, ``{Poisoning-Assisted
  Property Inference Attack Against Federated Learning},'' \emph{IEEE
  Transactions on Dependable and Secure Computing}, 2022.

\bibitem{tan2023general}
C.~Tan, Y.~Cao, S.~Li, and M.~Yoshikawa, ``General or specific? investigating
  effective privacy protection in federated learning for speech emotion
  recognition,'' in \emph{ICASSP 2023-2023 IEEE International Conference on
  Acoustics, Speech and Signal Processing (ICASSP)}.\hskip 1em plus 0.5em minus
  0.4em\relax IEEE, 2023, pp. 1--5.

\bibitem{Truex2019-iz}
S.~Truex, L.~Liu, M.~E. Gursoy, L.~Yu, and W.~Wei, ``{Demystifying Membership
  Inference Attacks in Machine Learning as a Service},'' \emph{IEEE Trans.
  Serv. Comput.}, pp. 1--1, 2019.

\bibitem{nasr19}
\BIBentryALTinterwordspacing
M.~Nasr, R.~Shokri, and A.~Houmansadr, ``{Comprehensive Privacy Analysis of
  Deep Learning: Passive and Active White-box Inference Attacks against
  Centralized and Federated Learning},'' in \emph{2019 {IEEE} Symposium on
  Security and Privacy, {SP} 2019, San Francisco, CA, USA, May 19-23,
  2019}.\hskip 1em plus 0.5em minus 0.4em\relax {IEEE}, 2019, pp. 739--753.
  [Online]. Available: \url{https://doi.org/10.1109/SP.2019.00065}
\BIBentrySTDinterwordspacing

\bibitem{shokri17}
R.~{Shokri}, M.~{Stronati}, C.~{Song}, and V.~{Shmatikov}, ``{Membership
  Inference Attacks Against Machine Learning Models},'' in \emph{2017 IEEE
  Symposium on Security and Privacy (SP)}, 2017, pp. 3--18.

\bibitem{ye2022enhanced}
J.~Ye, A.~Maddi, S.~K. Murakonda, V.~Bindschaedler, and R.~Shokri, ``{Enhanced
  Membership Inference Attacks against Machine Learning Models},'' in
  \emph{Proceedings of the 2022 ACM SIGSAC Conference on Computer and
  Communications Security}, 2022, pp. 3093--3106.

\bibitem{suri2023dissecting}
A.~Suri, Y.~Lu, Y.~Chen, and D.~Evans, ``{Dissecting Distribution Inference},''
  in \emph{IEEE Conference on Secure and Trustworthy Machine Learning (SaTML)},
  2023.

\bibitem{ruder2016overview}
S.~Ruder, ``An overview of gradient descent optimization algorithms,''
  \emph{arXiv preprint arXiv:1609.04747}, 2016.

\bibitem{wanggeneralizing}
J.~Wang, C.~Lan, C.~Liu, Y.~Ouyang, and T.~Qin, ``{Generalizing to Unseen
  Domains: A Survey on Domain Generalization},'' \emph{Proceedings of the
  Thirteenth International Joint Conference on Artificial Intelligence (IJCAI),
  Survey Track}, 2021.

\bibitem{hardt2016train}
M.~Hardt, B.~Recht, and Y.~Singer, ``{Train faster, generalize better:
  Stability of stochastic gradient descent},'' in \emph{International
  conference on machine learning}.\hskip 1em plus 0.5em minus 0.4em\relax PMLR,
  2016, pp. 1225--1234.

\bibitem{caldas18}
S.~Caldas, P.~Wu, T.~Li, J.~Kone{\v{c}}n{\'y}, H.~B. McMahan, V.~Smith, and
  A.~Talwalkar, ``{LEAF: A Benchmark for Federated Settings},'' \emph{CoRR},
  vol. abs/1812.01097, 2018.

\bibitem{deng2012mnist}
L.~Deng, ``{The MNIST Database of Handwritten Digit Images for Machine Learning
  Research},'' \emph{IEEE signal processing magazine}, vol.~29, no.~6, pp.
  141--142, 2012.

\bibitem{kingma2014adam}
D.~P. Kingma and J.~Ba, ``{Adam: A Method for Stochastic Optimization },''
  \emph{arXiv preprint arXiv:1412.6980}, 2014.

\bibitem{warner1965randomized}
S.~L. Warner, ``{Randomized Response: A Survey Technique for Eliminating
  Evasive Answer Bias},'' \emph{Journal of the American Statistical
  Association}, vol.~60, no. 309, pp. 63--69, 1965.

\bibitem{evfimievski2003limiting}
A.~Evfimievski, J.~Gehrke, and R.~Srikant, ``{Limiting Privacy Breaches in
  Privacy Preserving Data Mining},'' in \emph{Proceedings of the twenty-second
  ACM SIGMOD-SIGACT-SIGART symposium on Principles of database systems}, 2003,
  pp. 211--222.

\bibitem{kasiviswanathan08}
\BIBentryALTinterwordspacing
S.~P. Kasiviswanathan, H.~K. Lee, K.~Nissim, S.~Raskhodnikova, and A.~D. Smith,
  ``What can we learn privately?'' \emph{CoRR}, vol. abs/0803.0924, 2008.
  [Online]. Available: \url{http://arxiv.org/abs/0803.0924}
\BIBentrySTDinterwordspacing

\bibitem{papyan2020prevalence}
V.~Papyan, X.~Y. Han, and D.~L. Donoho, ``Prevalence of neural collapse during
  the terminal phase of deep learning training,'' \emph{Proceedings of the
  National Academy of Sciences}, vol. 117, no.~40, 2020.

\end{thebibliography}

\onecolumn
\clearpage
\appendix
\section{Appendix}
\subsection{Measuring Attack Success}
\label{sec:specific_configs_synthetic}

\begin{figure*}[bh]
\centering
\begin{subfigure}[b]{0.328\textwidth}
    \centering
    \includegraphics[width=\textwidth]{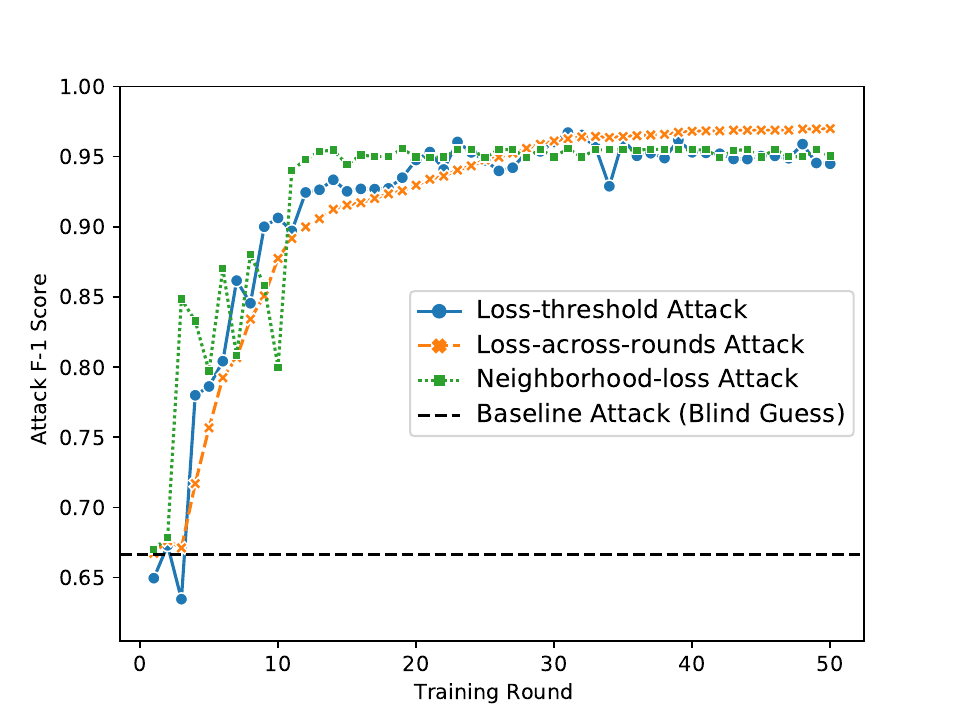}
    \caption{Config \textit{A}. Final model test accuracy: 99.19\%}
    \label{fig:all_att_1}
\end{subfigure}
\begin{subfigure}[b]{0.328\textwidth}
    \centering
    \includegraphics[width=\textwidth]{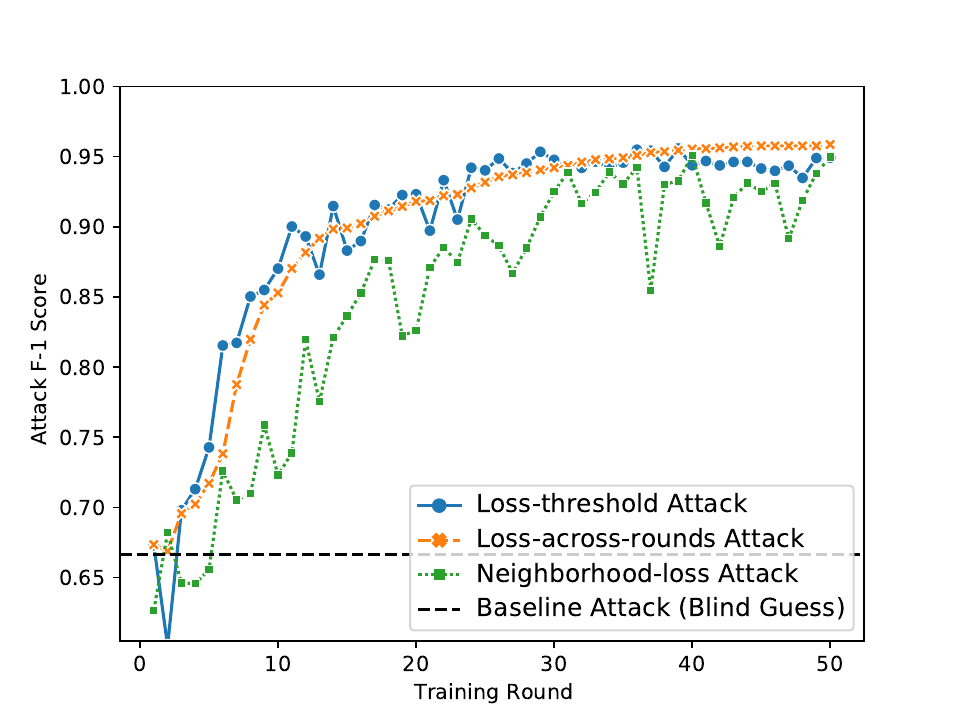}
    \caption{Config \textit{B}. Final model test accuracy: 99.82\%}
    \label{fig:all_att_2}
\end{subfigure}
\begin{subfigure}[b]{0.328\textwidth}
    \centering
    \includegraphics[width=\textwidth]{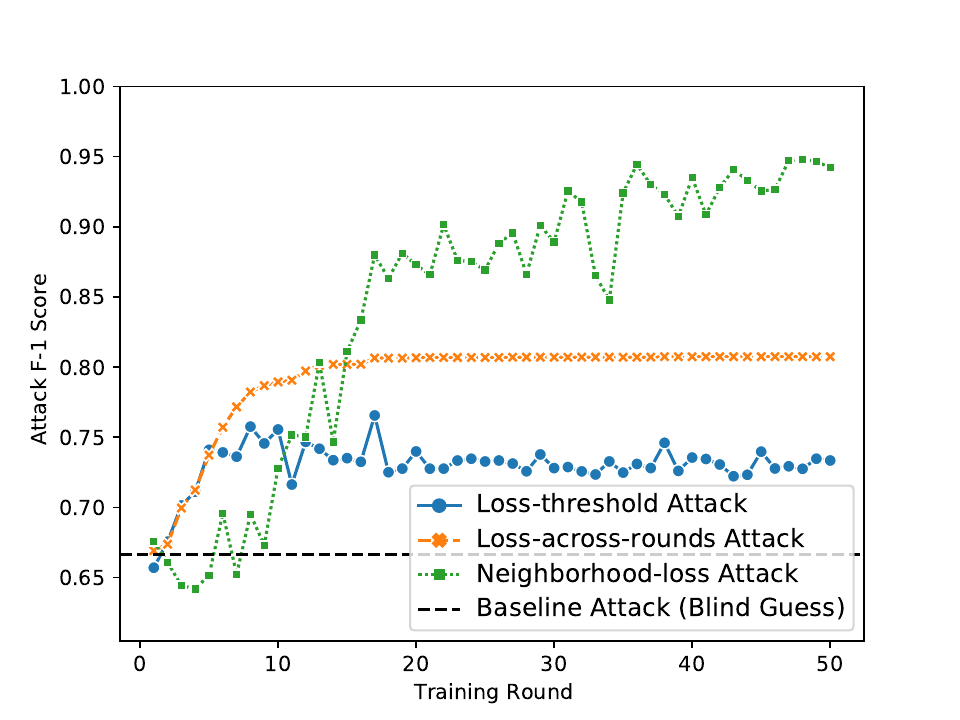}
    \caption{Config \textit{C}. Final model test accuracy: 50.35\%}
    \label{fig:all_att_3}
\end{subfigure}
\begin{subfigure}[b]{0.328\textwidth}
    \centering
    \includegraphics[width=\textwidth]{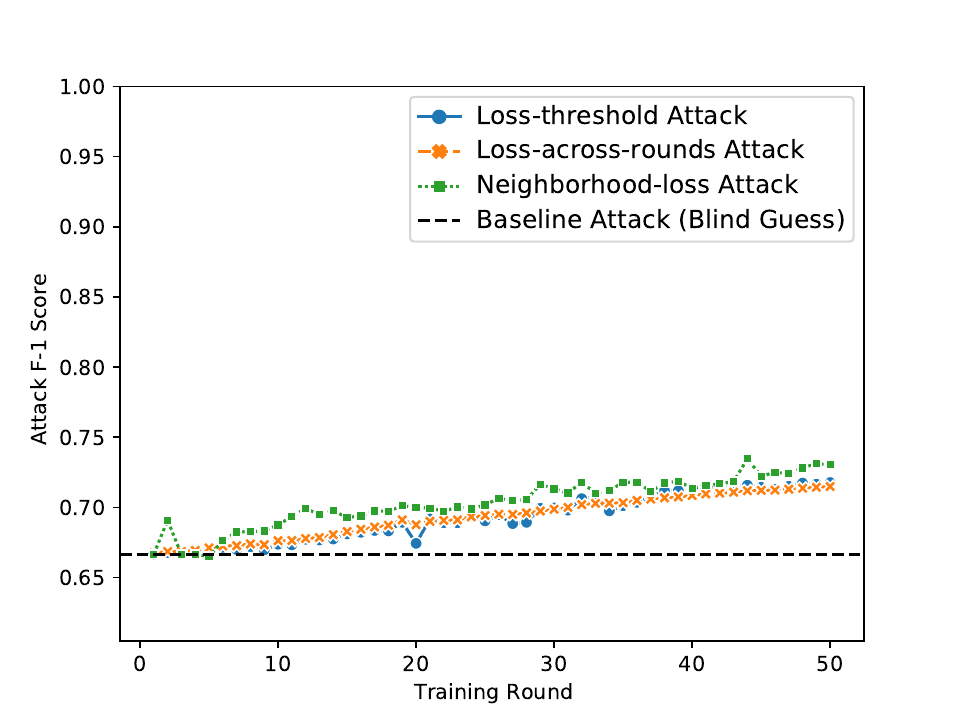}
    \caption{Config \textit{D}. Final model test accuracy: 51.26\%}
    \label{fig:all_att_4}
\end{subfigure}
\hfill
\begin{subfigure}[b]{0.328\textwidth}
    \centering
    \includegraphics[width=\textwidth]{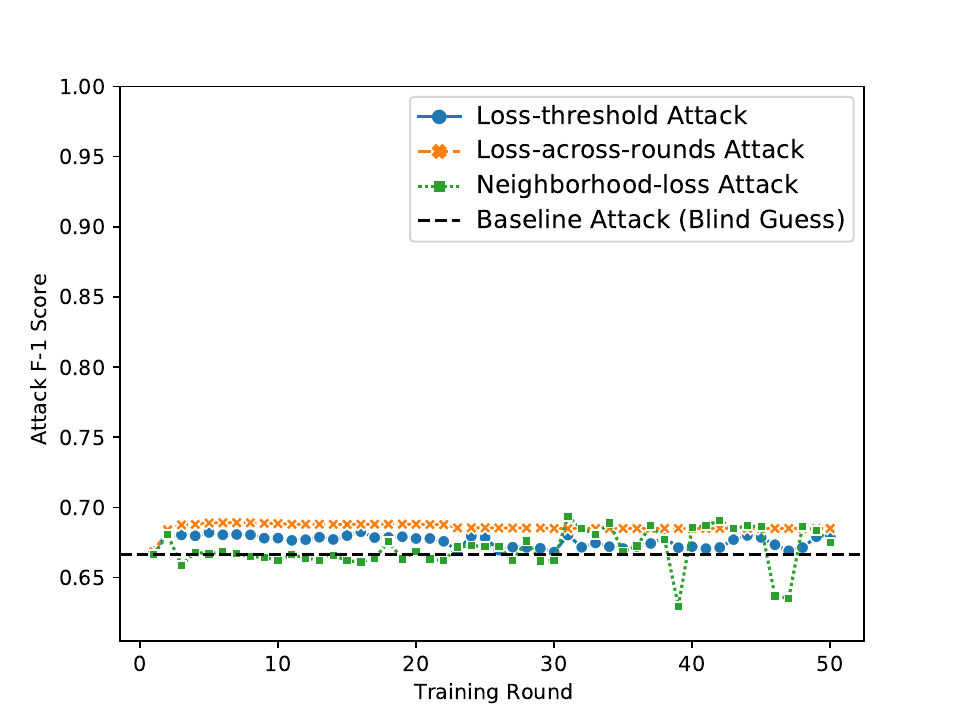}
    \caption{Config \textit{E}. Final model test accuracy: 99.85\%}
    \label{fig:all_att_5}
\end{subfigure}
\begin{subfigure}[b]{0.328\textwidth}
    \centering
    \includegraphics[width=\textwidth]{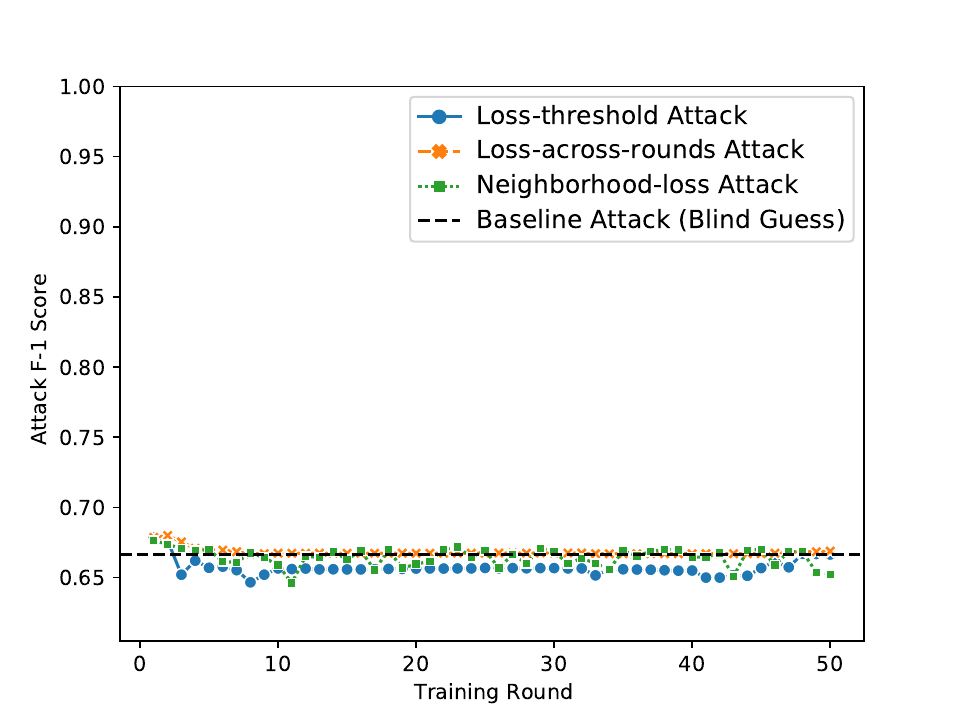}
    \caption{Config \textit{F}. Final model test accuracy: 65.45\%}
    \label{fig:all_att_6}
\end{subfigure}
\hfill
\caption{Attack F-1 Scores for configurations with varying final test accuracies.
For these experiments, we also tried a variant that analyzes loss in neighborhood of data to make predictions (Neighborhood Loss Attack).
We observe a full spectrum of attack success for different configurations. From configurations in which the attacks are highly accurate (\protect{\subref{fig:all_att_1}}, \protect{\subref{fig:all_att_2}}) to the cases where there is close to little or no leakage when models (\protect{\subref{fig:all_att_5}}, \protect{\subref{fig:all_att_6}}). We see that in general, there is a strong correlation between the effectiveness of the three proposed attack, but there doesn't seem to be a strong correlation between attack F-1 and model test accuracy.}
\label{fig:all_attacks}
\end{figure*}

We first present results on six example configurations that cover the full range of attack success. These configurations also represent a good variety in the various environmental variables like sampling mechanism, data dimensionality, and model capacity. The exact configuration parameters are given in Table~\ref{tab:all_attack_configs}. Results for all three attacks and configurations are plotted in Figure~\ref{fig:all_attacks}.


\subsection{Mitigation}
\label{sec:mitigate}

For our synthetic dataset experiments, from the 720 configurations described earlier, we select the most vulnerable ones, and train models on them with DP at $\epsilon=2.0$ and $\delta=10^{-5}$ for all three privacy granularities (item, user, and subject level\cite{marathesubject}). We train these models for 20 rounds, with a mini-batch size of $20$, and use $\sigma=1.8346$.  

We first assess how well the three attacks behave on a few example configurations (Section~\ref{sec:exploratory_results}). For this purpose, we look at a few representative configurations: Config \textit{A}, Config \textit{B}, and Config \textit{C}, and examine the full range of attack success. These configurations are selected to have a good variety in the various environmental variables like sampling mechanism, data dimensionality, and model capacity. The exact configuration parameters are given in Table~\ref{tab:all_attack_configs}. 
\begin{table}[h]
    \centering
    \begin{tabular}{c c c l c}
        \toprule
        Configuration & Data & Sampling & Model Hidden & Sub/User\\
        \midrule
        Config \textit{A} & 1000 & Dirichlet & [256, 64, 16, 4] & 10\\
        Config \textit{B} & 1000 & Dirichlet & [128,32,8] & 10 \\
        Config \textit{C} & 1000 & Normal & [8] & 10\\
        Config \textit{D} & 1000 & Normal & [2] & 500\\
        Config \textit{E} & 2 & Normal & [128, 32, 8] & 100\\
        Config \textit{F} & 2 & Normal & [2] & 10\\
         \bottomrule
    \end{tabular}
    \caption{Experiment parameters for the configurations described in Section~\ref{sec:exploratory_results}. Sub/User is the number of subjects per user. All of these configurations correspond to 10000 items per user, with 10 users for all but Configs \textit{D} and \textit{E}, which have 100 users.}
    \label{tab:all_attack_configs}
\end{table}

\begin{table}[h]
    \centering
    \begin{tabular}{c c c c c}
         \toprule
         Metric & FL & Item & User & Subject \\
         \midrule
         \multicolumn{5}{c}{\textbf{Synthetic Dataset Config \textit{A}}} \\
         \midrule
         Model Accuracy & $.9919$ & $.7945$ & $.7290$ & $.6368$\\
         \hline
         Accuracy & $.93\pm.01$ & $.66\pm.04$ & $.59\pm.02$ & $.58\pm.05$ \\
         Precision & $.89\pm.02$ & $.61\pm.04$ & $.55\pm.02$ & $.55\pm.03$\\
         Recall & $.98\pm.02$ & $.93\pm.06$ & $.98\pm.02$ & $.89\pm.05$ \\
         $F_1$ Score & $.93\pm.01$ & $.74\pm.01$ & $.71\pm.01$ & $.68\pm.02$\\
         \midrule
         \multicolumn{5}{c}{\textbf{Synthetic Dataset Config \textit{C}}} \\
         \midrule
         Model Accuracy & $.5035$ & $.5085$ & $.5018$ & $.5075$\\
         \hline
         Accuracy & $.78\pm.02$ & $.52\pm.04$ & $.50\pm.01$ & $.52\pm.03$\\
         Precision & $.73\pm.04$ & $.51\pm.02$ & $.50\pm.00$ & $.51\pm.02$\\
         Recall & $.91\pm.05$ & $.97\pm.06$ & $1.0\pm.00$ & $.98\pm.03$\\
         $F_1$ Score & $.81\pm.02$ & $.67\pm.00$ & $.67\pm.00$ & $.67\pm.01$\\
         \midrule
         \multicolumn{5}{c}{\textbf{Synthetic Dataset Config \textit{F}}} \\
         \midrule
         Model Accuracy & $.6545$ & $.6291$ & $.8358$ & $.6383$ \\
         \hline
         Accuracy & $.53\pm.04$ & $.53\pm.04$ & $.52\pm.03$ & $.50\pm.01$ \\
         Precision & $.51\pm.01$ & $.52\pm.02$ & $.51\pm.02$ & $.50\pm.01$\\
         Recall & $.98\pm.03$ & $.97\pm.04$ & $.98\pm.03$ & $1.0\pm.01$ \\
         $F_1$ Score & $.67\pm.00$ & $.68\pm.01$ & $.67\pm.01$ & $.67\pm.00$\\
        \bottomrule
    \end{tabular}
    \captionsetup{justification=centering}
    \caption{Model accuracies and attack metrics (accuracy, precision, recall, $F_1$ score) under different DP granularities while using the \textit{Loss-Threshold Attack}, using MLPs on the Synthetic Dataset (Section~\ref{sec:synthetic_data}). DP across all granularities provide near-perfect robustness against attacks, albeit at the cost of huge drops in model accuracy.}
    \label{table:dp_synthetic_results}
\end{table}

Table~\ref{table:dp_synthetic_results} depicts the model accuracy and attack efficacy (accuracy, precision, recall, and $F_1$ Score) when DP is introduced while training models for three\footnote{Since the DP experiments are computationally expensive and many of the six configurations are similar, we report results with DP for only three of them.} of our representative configurations introduced in Section~\ref{sec:exploratory_results}. The FL column for all three configurations shows the models covering different ranges of performance. Models with high risk configurations such as Config $A$ are susceptible to subject membership inference attacks. Interestingly, poorly performing models (from Config $C$) can also be vulnerable to such attacks. Configurations like Config $F$ have low subject membership inference risk even without any DP, further reinforcing our observation of some configurations being significantly easier/harder to attack than others.   

Results for all the configurations show that DP at all granularities provides non-trivial robustness against subject membership inference attacks.  However, DP enforcement generally leads to performance degradation of the trained model as expected.  The progressive degradation from item- to user- to subject-level DP algorithms is more evident in the high performing model of Config $A$, which is intuitive, given the increasing strictness of the privacy guarantees.  Since Config $C$ and $F$'s models perform relatively poorly to begin with, the DP related noise injection does not seem to significantly affect model performance (we are investigating the anomalous performance of Config $F$ with user-level DP).

\subsection{Additional Figures}
\label{app:exps}

\begin{figure*}
\centering
\begin{subfigure}[b]{0.49\textwidth}
    \centering
    \includegraphics[width=\textwidth]{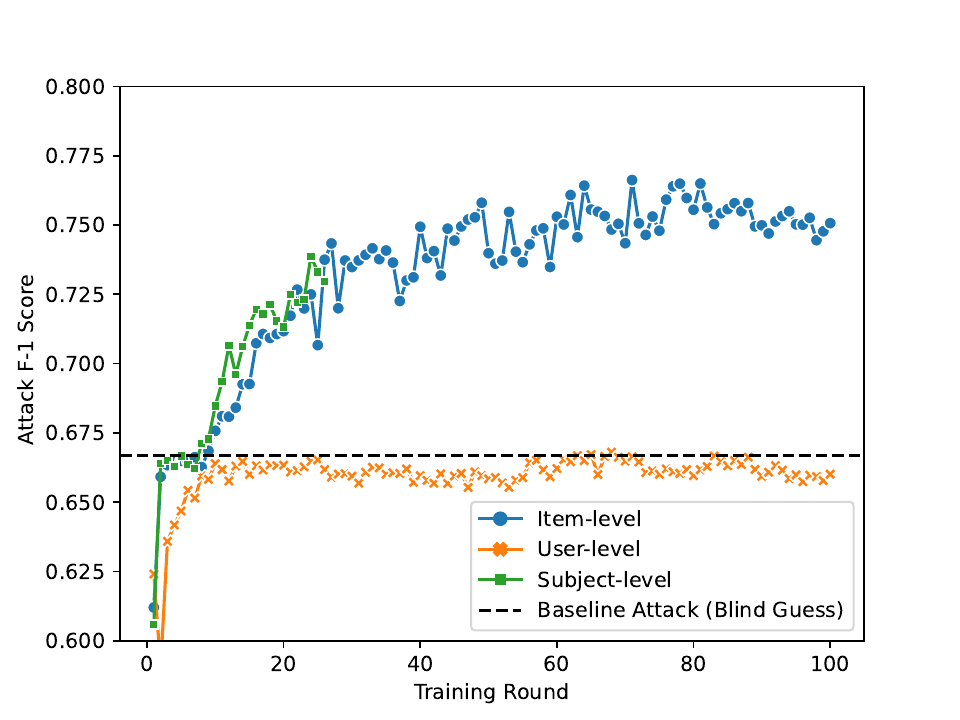}
    \caption{Item-based Subject Membership}
    \label{fig:dp_training_logs_item}
\end{subfigure}
\hfill
\begin{subfigure}[b]{0.49\textwidth}
    \centering
    \includegraphics[width=\textwidth]{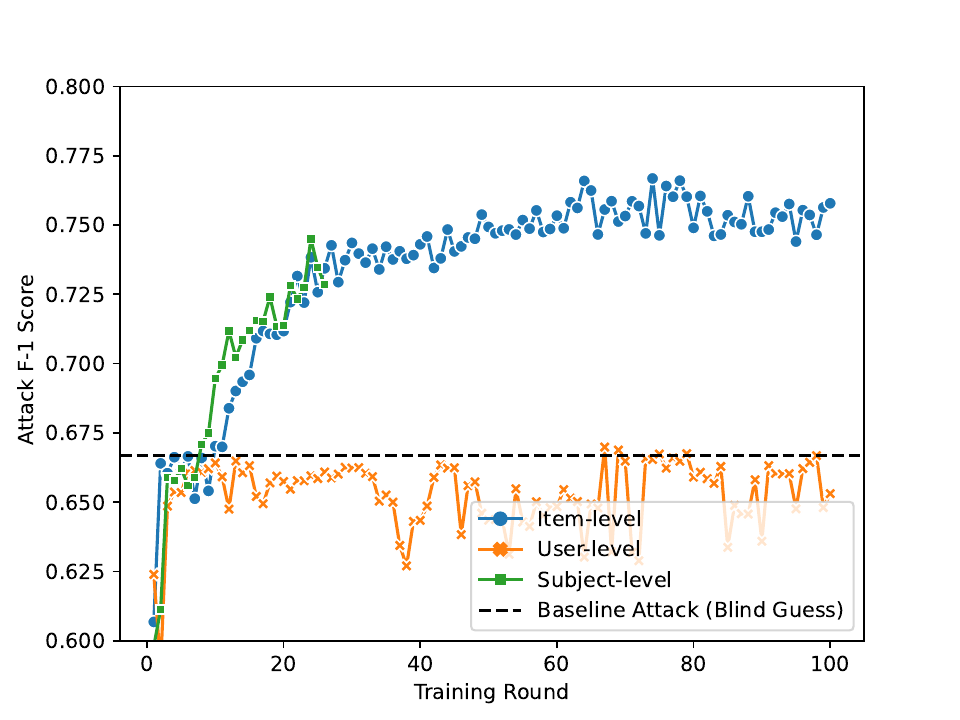}
    \caption{Distribution-based Subject Membership}
    \label{fig:dp_training_logs_dist}
\end{subfigure}
\hfill
\caption{Attack F-1 Score across training rounds inferring subject membership, for Item-based (\protect{\subref{fig:dp_training_logs_item}}) and Distribution-based (\protect{\subref{fig:dp_training_logs_dist}}), for various DP granularities in FL. User-level DP completely eliminates risk, although at the cost of a huge dent in task performance. Subject-level DP, as expected, leads to lower final inference risk.}
\label{fig:dp_training_logs}
\end{figure*}

\begin{figure*}
\centering
\begin{subfigure}[b]{0.49\textwidth}
    \centering
    \includegraphics[width=\textwidth]{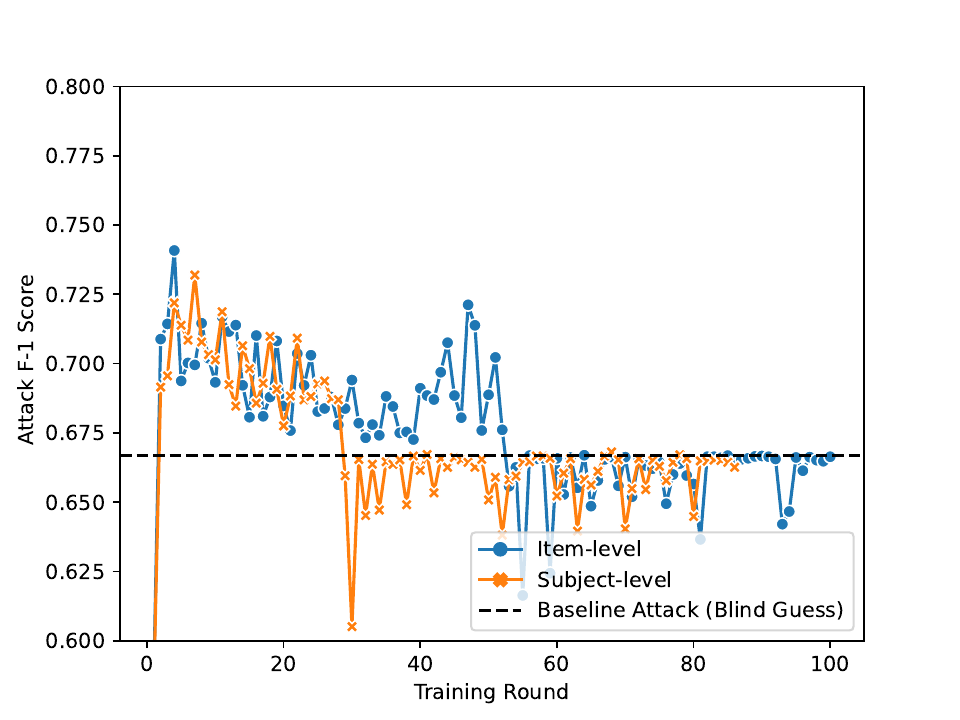}
    \caption{Item-based Subject Membership}
    \label{fig:dp_training_logs_standalone_item}
\end{subfigure}
\hfill
\begin{subfigure}[b]{0.49\textwidth}
    \centering
    \includegraphics[width=\textwidth]{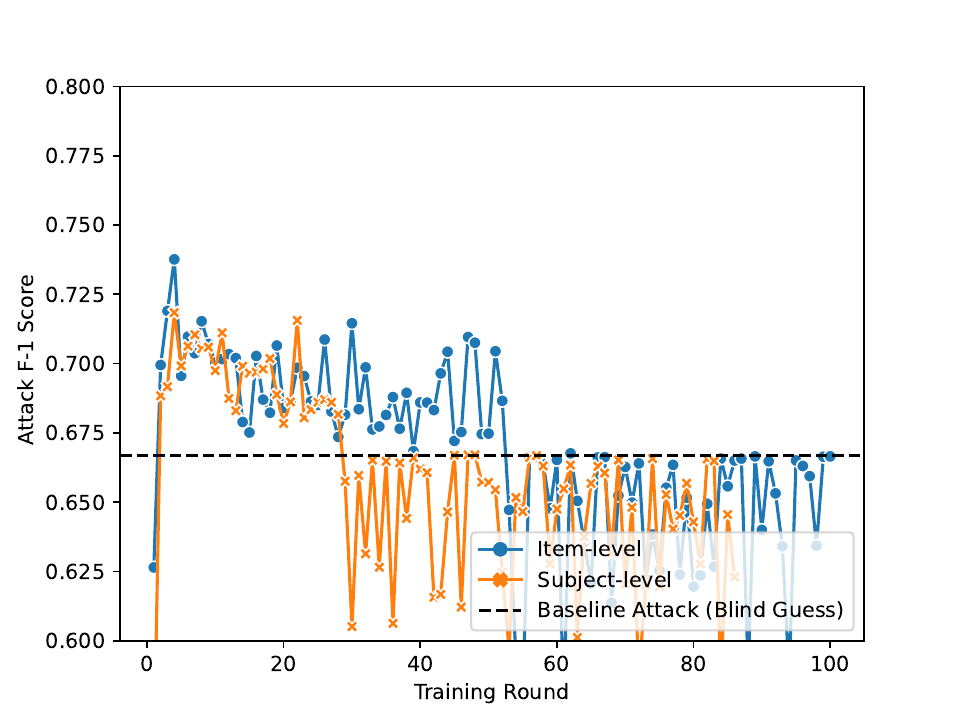}
    \caption{Distribution-based Subject Membership}
    \label{fig:dp_training_logs_standalone_dist}
\end{subfigure}
\hfill
\caption{Attack F-1 Score across training rounds inferring subject membership, for Item-based (\protect{\subref{fig:dp_training_logs_standalone_item}}) and Distribution-based (\protect{\subref{fig:dp_training_logs_standalone_item}}), for various DP granularities in standard training. Both notions of DP eliminate inference risk.}
\label{fig:dp_training_logs_standalone}
\end{figure*}

\begin{figure*}
\centering
\begin{subfigure}[b]{0.49\textwidth}
    \centering
    \includegraphics[width=\textwidth]{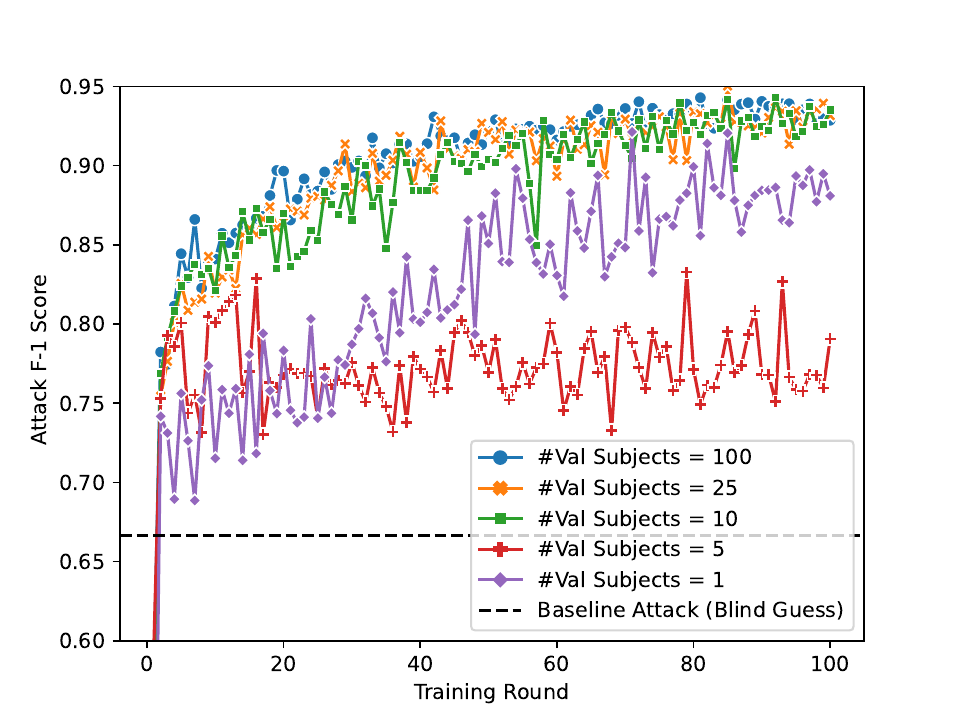}
    \caption{Normal Training}
\end{subfigure}
\hfill
\begin{subfigure}[b]{0.49\textwidth}
    \centering
    \includegraphics[width=\textwidth]{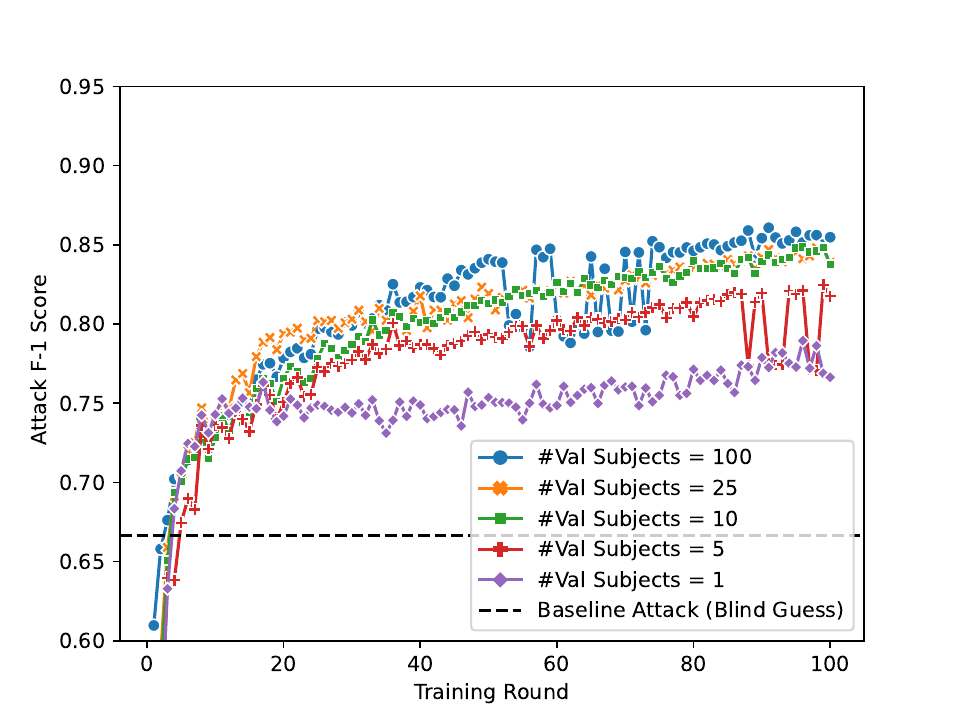}
    \caption{Federated Training}
    \label{fig:num_subject_val_item}
\end{subfigure}
\hfill
\caption{Attack F-1 Score across training rounds inferring subject membership using our attack with Item-based Subject Membership, for normal training (a) and FL (b), while varying the number of subjects in-set used for validation.}
\label{fig:two_attacks_and_dp_grand_item}
\end{figure*}

\begin{figure*}
\centering
\begin{subfigure}[b]{0.49\textwidth}
    \centering
    \includegraphics[width=\textwidth]{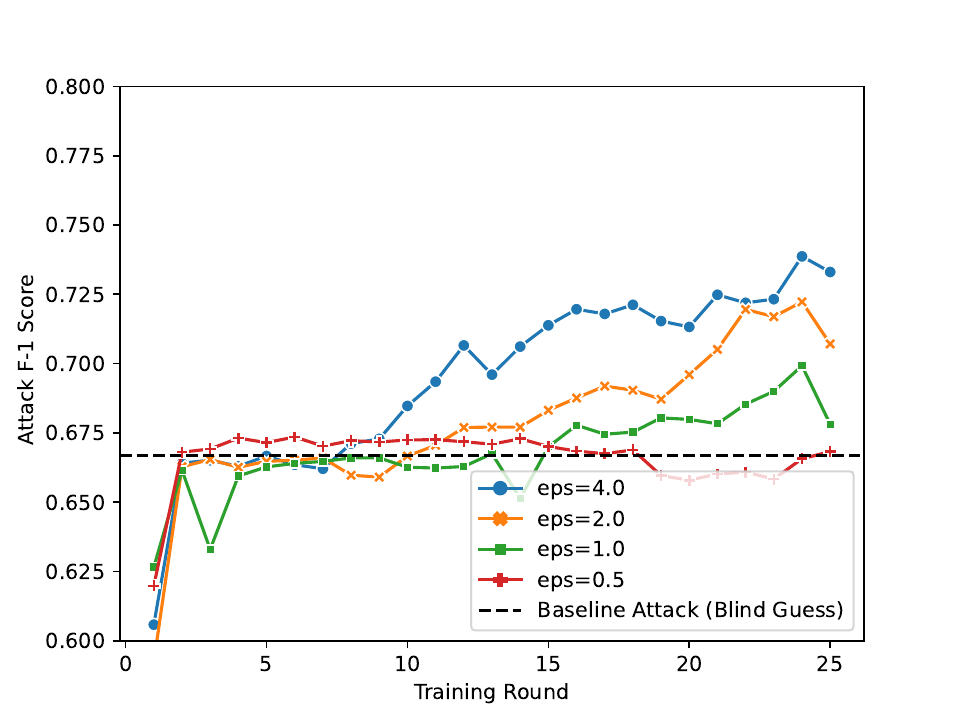}
    \caption{Item-based Subject Membership}
    \label{fig:fl_item_vary_eps_item}
\end{subfigure}
\hfill
\begin{subfigure}[b]{0.49\textwidth}
    \centering
    \includegraphics[width=\textwidth]{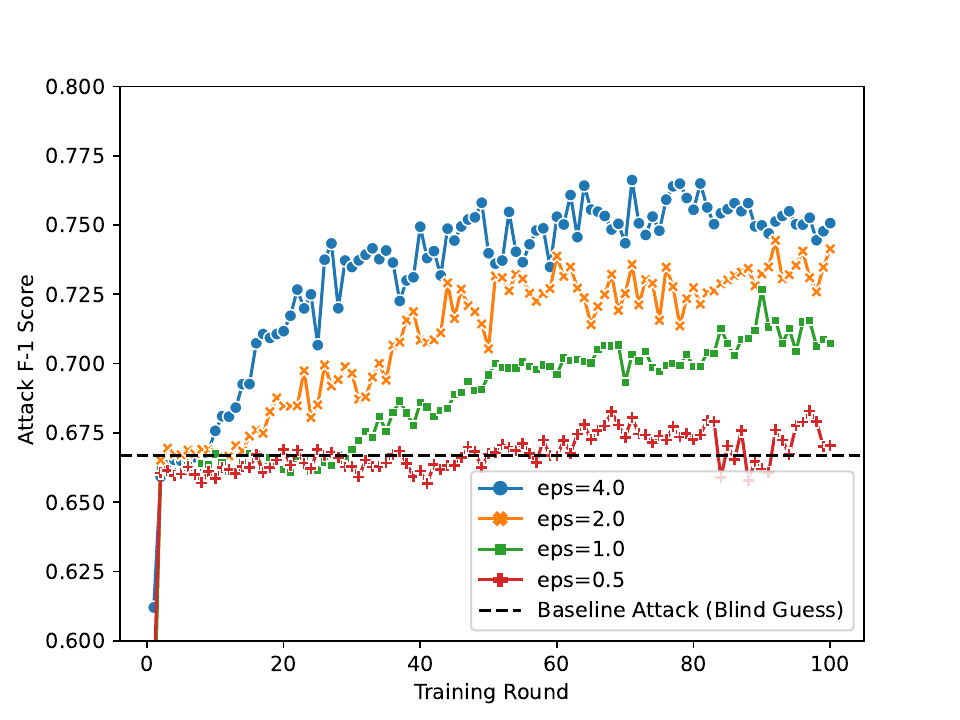}
    \caption{Distribution-based Subject Membership}
    \label{fig:fl_sub_vary_eps_item}
\end{subfigure}
\hfill
\caption{Attack F-1 Score across training rounds inferring subject membership using our attack with Item-based Subject Membership, for Item-level DP (\protect{\subref{fig:fl_item_vary_eps_item}}) and Subject-level DP (\protect{\subref{fig:fl_sub_vary_eps_item}}) with varying levels of protection ($\epsilon$). Both variants of the attacks are equivalent in potency, irrespective of the granularity of DP used for protection.}
\label{fig:dp_epsilons_item}
\end{figure*}

\begin{figure*}
\centering
    \includegraphics[width=0.49\textwidth]{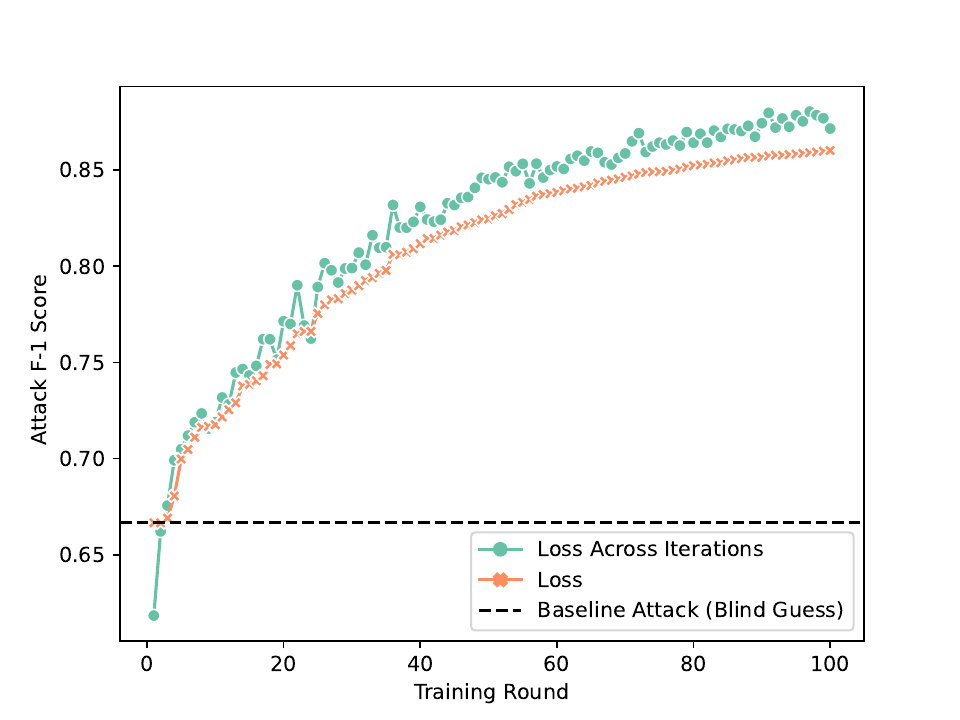}
    \caption{Attack F-1 Score across training rounds inferring subject membership using our attack with Item-based Subject Membership for the two proposed attacks (Loss-Threshold and Loss-Across-Rounds)}
    \label{fig:both_attacks}
\end{figure*}


\end{document}